\documentclass[10pt,twocolumn,letterpaper]{article}

\usepackage{iccv}
\usepackage{times}
\usepackage{epsfig}
\usepackage{graphicx}
\usepackage{amsmath}
\usepackage{amssymb}

\DeclareMathOperator*{\argmax}{argmax}

\usepackage{tabularx} %Nguyen
\newcolumntype{Y}{>{\centering\arraybackslash}X}
\usepackage{adjustbox} %Nguyen

\usepackage[numbers]{natbib}
\usepackage{multirow}

\usepackage[breaklinks=true,bookmarks=false]{hyperref}

\iccvfinalcopy % *** Uncomment this line for the final submission

\def\iccvPaperID{****} % *** Enter the ICCV Paper ID here

\ificcvfinal\fi

\begin{document}

%%%%%%%%% TITLE
\title{Anomaly Detection in Video Sequence with Appearance-Motion Correspondence}

\author{Trong-Nguyen Nguyen, Jean Meunier\\
DIRO, University of Montreal\\
{\tt\small \{nguyetn, meunier\}@iro.umontreal.ca}
}

\maketitle
\ificcvfinal\thispagestyle{empty}\fi

%%%%%%%%% ABSTRACT
\begin{abstract}
Anomaly detection in surveillance videos is currently a challenge because of the diversity of possible events. We propose a deep convolutional neural network (CNN) that addresses this problem by learning a correspondence between common object appearances (\eg pedestrian, background, tree, etc.) and their associated motions. Our model is designed as a combination of a reconstruction network and an image translation model that share the same encoder. The former sub-network determines the most significant structures that appear in video frames and the latter one attempts to associate motion templates to such structures. The training stage is performed using only videos of normal events and the model is then capable to estimate frame-level scores for an unknown input. The experiments on 6 benchmark datasets demonstrate the competitive performance of the proposed approach with respect to state-of-the-art methods.
\end{abstract}

%%%%%%%%% BODY TEXT
\section{Introduction}

Anomaly detection in video sequences is a necessary functionality for surveillance systems. Because abnormal events rarely occur in real-world videos, this task is significantly time-consuming and may require a large amount of resource (\eg people) to perform manual checking. A method than can automatically determine potential frames of anomalous events is thus crucial.

%\colorbox[rgb]{1,1,0}{TO DO LATER}

Our model is a combination of a convolutional auto-encoder (Conv-AE) and a U-Net with skip connections~\cite{Olaf2015UNet} that share the same encoder sub-network. Other related works employed either an AE or a U-Net to perform the anomaly detection in different ways. Hasan~\etal~\cite{Hasan2016Learning} estimate regularity score for frames in video sequences according to reconstruction models. Their two AEs (with and without convolutional layers) work on two different inputs: hand-crafted features (HOG and HOF with trajectory-based properties~\cite{Wang2013Action}) and concatenation of 10 consecutive frames along the temporal axis. The reconstruction error is used to indicate their regularity score. Unlike that work, the input of our Conv-AE is a single frame and the temporal factor is considered in the other stream via U-Net. The purpose of our Conv-AE is to learn only regular appearance structures.

On the contrary, Ravanbakhsh~\etal~\cite{Ravanbakhsh2017Abnormal} employ the U-Net structure proposed in~\cite{Isola2017Image} to translate an input from video frame to a corresponding optical flow and vice versa. We argue that the use of two CNNs with the same structure may be redundant and an appropriate modification and/or combination would improve the model ability. Compared with~\cite{Ravanbakhsh2017Abnormal}, our network keeps the stream translating a video frame to an optical flow (but using our proposed structure instead of~\cite{Isola2017Image}) while replaces the other U-Net by a Conv-AE that shares the encoding flow.

Inspired by the good performance of the video prediction model in~\cite{Mathieu2015Deep}, Liu~\etal~\cite{Wen2018Future} present a model that uses a U-Net structure to predict a frame from a number of recent ones and then estimates the corresponding optical flow. The model is optimized according to the difference between the outputted and original versions of video frame as well as the optical flow together with an adversarial loss. Our work also predicts an optical flow but directly from a single frame in order to determine the association between a scene appearance and its typical motion. Since a fixed procedure of optical flow estimation (FlowNet~\cite{Dosovitskiy2015FlowNet}) is embedded inside the network in~\cite{Wen2018Future}, the selection of such method is thus limited because the estimator has to be fully differentiable to perform an end-to-end training. Our model, however, has a stream that directly estimates a mapping from input frame to optical flow. We only use a pretrained estimator for ground truth calculation and the model signal does not propagate through it during the training as well as inference stages.

Our main contributions are summarized as follows:
\begin{itemize}
	\item We design a CNN that combines a Conv-AE and a U-Net, in which each stream has its own contribution for the task of detecting anomalous frames. The model can be trained end-to-end.
	\item We integrate an Inception module modified from~\cite{Szegedy2016Rethinking} right after the input layer to reduce the effect of network's depth since this depth is considered as a hyperparameter that requires a careful selection.
	\item We propose a patch-based scheme estimating frame-level normality score that reduces the effect of noise which appears in the model outputs.
	\item Experiments on 6 benchmark datasets demonstrate the potential of our model with competitive performance compared with state-of-the-art methods. We also provide discussions for these datasets that should be useful for future works.
\end{itemize}

The remainder of this paper is organized as follows: a summary of related studies is given in Section~\ref{sec:related}; Section~\ref{sec:method} describes the details of our method; experiments and discussions for the 6 benchmark datasets are presented in Section~\ref{sec:experiment}; and Section~\ref{sec:conclusion} concludes this work.

\section{Related work}\label{sec:related}
We briefly describe the principal categories that lead to very different approaches for anomaly detection in video.

\subsection{Trajectory}
The diversity of possible anomalous events is the main challenge of the anomaly detection problem. Some researchers simplify this issue by explicitly specifying anomalies (\eg~\cite{Sultani2018}) or particular relevant attributes that can be used effectively for anomaly detection, in which the most common one is motion trajectory. These studies aim to learn patterns of object trajectories determined from normal events~\cite{Medioni2001Event,Basharat2008Learning,Piciarelli2008Trajectory,Zhang2009Learning}. There are four main stages in the methodology including object detection, tracking, trajectory-based feature extraction and classification/detection. The advantages of methods in this category are the simple implementation and fast execution. However, their effectiveness may significantly degrade when working on videos with cluttered background since the trajectory determination depends on the result of object detection and tracking. Moreover, trajectory anomalies do not cover the whole spectrum of anomalies in video surveillance.

\subsection{Sparse coding}
Instead of explicitly defining and estimating specific anomaly attributes, other researchers consider an input sequence of frames as a collection of small 3D patches. Concretely, a number of consecutive frames are concatenated along the temporal axis and then split into same-size 3D patches according to a window sliding on the image plane. In the inference stage, each 3D patch extracted from unknown inputs is represented as a sparse combination of training samples of normal events. The reconstruction error is considered as the score supporting the final decision. Such sparsity-based methods have achieved state-of-the-art performances~\cite{Cong2011Sparse,Zhao2011Online}. The main drawback is the high computational cost in finding combination coefficients due to sparse representation. Some studies thus attempt to reduce the complexity by modifying the learning algorithms and/or data structures~\cite{Lu2013Abnormal,Luo2017A}. Beside window-based split, 3D patches are also determined using keypoint detectors~\cite{Cheng2015Video} while other researchers attempt to learn the relation between training patches according to their distribution~\cite{Mahadevan2010Anomaly} or graph-based representation~\cite{Kim2009Observe}.

\subsection{Deep learning}
Since deep learning models currently achieve top performance in a wide range of vision applications such as image classification~\cite{Krizhevsky2012ImageNet,Szegedy2015Going,He2016Deep}, object detection~\cite{Shaoqing2015Faster,He2017Mask} and image captioning~\cite{Johnson2016DenseCap,Karpathy2017Deep}, many CNNs have been proposed to deal with the problem of anomaly detection in videos. Typical structures of image reconstruction and translation are usually employed and the difference between their output and ground truth is used to indicate the frame-level score~\cite{Hasan2016Learning,Ravanbakhsh2017Abnormal,Wen2018Future}. Some researchers apply pretrained classification models (such as VGG~\cite{Simonyan2014Very}) to extract useful features from input videos~\cite{Sorina2017Deep,Ionescu2017Unmasking}. Results of object detection and/or foreground estimation are also used for the determination of anomalous events in~\cite{Hinami2017Joint,Xu2017Detecting}.

%\colorbox[rgb]{1,1,0}{TO DO LATER}

\section{Proposed method}\label{sec:method}
\begin{figure}[t]
\begin{center}
\begin{picture}(250,160)
\put(0,-10){\includegraphics[width=\columnwidth]{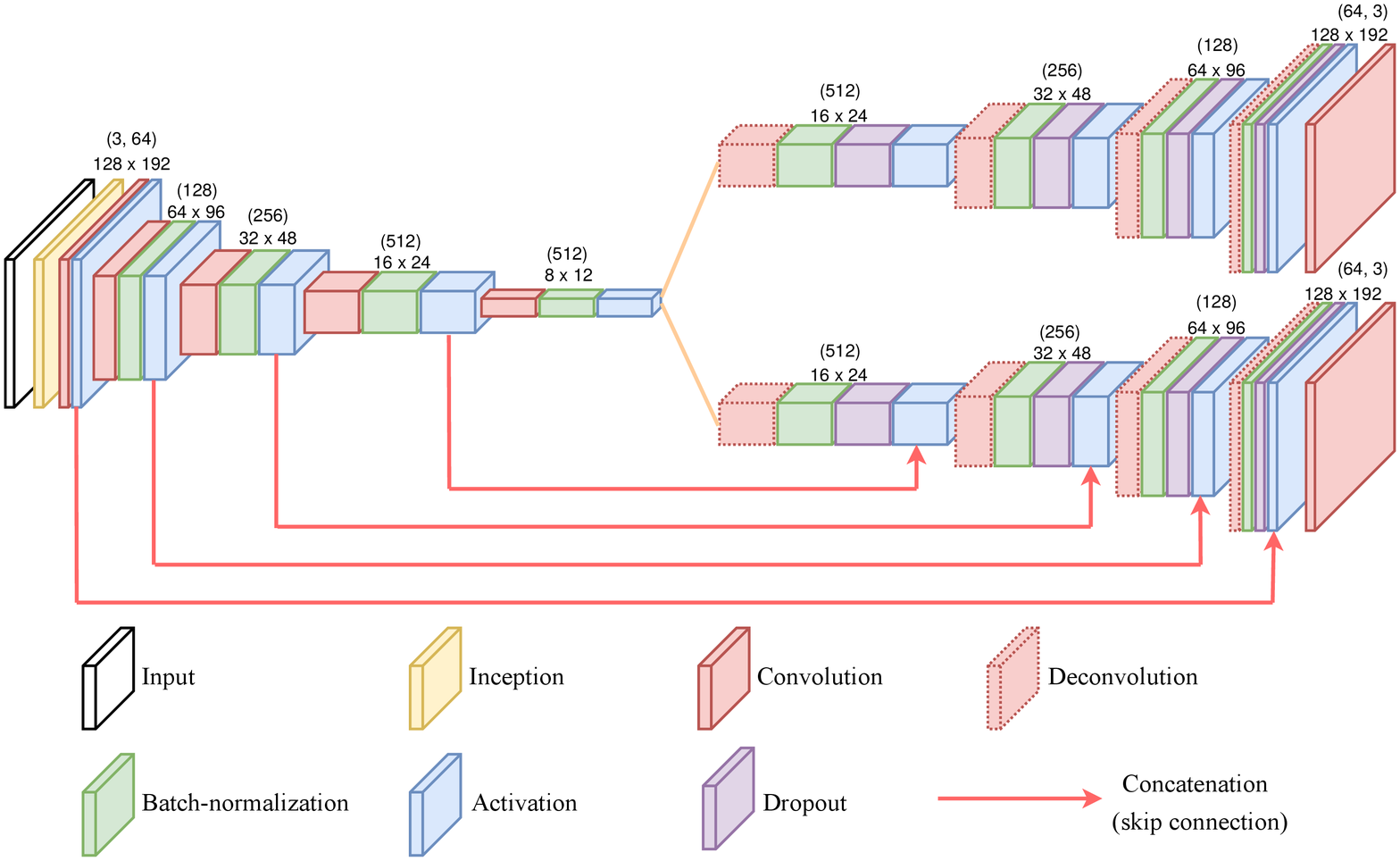}}
\put(40,115){\includegraphics[scale=0.5]{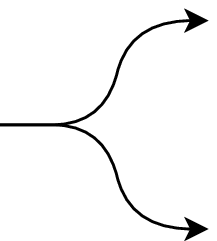}}
\put(8,119){\includegraphics[scale=0.35]{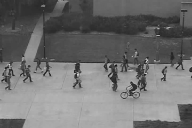}}
\put(70,134){\includegraphics[scale=0.35]{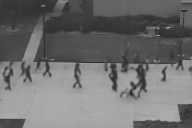}}
\put(70,104){\includegraphics[scale=0.35]{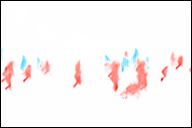}}
\put(25,150){$I_t$}\put(112,152){$\hat{I}_t$}\put(112,122){$\hat{F}_t$}
\end{picture}
\end{center}
\caption{Overview of our model structure together with the spatial resolution of feature maps in each block (\ie a sequence of layers with the same output shape). The number of channels corresponding to each layer in each block is also presented (in parentheses). The input and two output layers have the same size of $128\times192\times3$. There are three clusters of layers: common encoder (left), appearance decoder (top right) and motion decoder (bottom right). Each concatenation is performed along the channel axis right before operating the next deconvolution. The model input is a single video frame $I_t$ and the outputs from the two decoders are a reconstructed frame $\hat{I}_t$ and an optical flow $\hat{F}_t$ predicting the motion between $I_t$ and $I_{t+1}$. Best viewed in color.}
\label{fig:overview}
\end{figure}
An overview of our model is visualized in Figure~\ref{fig:overview}. The model includes two processing streams. The first one is performed via a Conv-AE to learn common appearance spatial structures in normal events. The second stream is to determine an association between each input pattern and its corresponding motion represented by an optical flow of 3 channels ($xy$ displacements and magnitude). The skip connections in U-Net are useful for image translation since it directly transforms low-level features (\eg edge, image patch) from original domains to the decoded ones. Such connections are not employed in the appearance stream because the network may let the input information go through these connections instead of emphasizing underlying attributes via the bottleneck.

Our model does not use any fully-connected layer, so it can theoretically work on images of any resolution. In order to simplify the model as well as make it be appropriate for possible further extensions, we fixed the size of input layer as $128\times192\times3$. The image size is set to a ratio of 1:1.5 instead of 1:1 as in related works (\eg~\cite{Hasan2016Learning,Sorina2017Deep,Wen2018Future}) in order to preserve the aspect of objects in surveillance videos.

\subsection{Inception module}
The Inception module was originally proposed to let a CNN decide its filter size (in a few layers) automatically~\cite{Szegedy2015Going}. A number of convolutional operations with various filter resolutions are performed in parallel and the obtained feature maps are then concatenated along the channel axis. The use of this module in our work can be explained under an alternative perspective as follows. The proposed network has an encoder-decoder structure with bottleneck. A very deep architecture may eliminate the features that are helpful for decoding. On the contrary, a shallow network takes the risk of missing high-level abstractions. Therefore, we apply an Inception module to let the model select its appropriate convolutional operations.

This work focuses on surveillance videos acquired from a fixed position. Given a convolutional layer with a predefined receptive field (\ie filter size) right after the input layer, the information abstraction would be different for the same object captured at various distances. This property is propagated for next layers, we thus expect the model to early determine low-level features by putting the Inception module right after the input layer. We remove the max-pooling in this module since the input is a regular video frame instead of a collection of feature maps. Our Inception module is modified from~\cite{Szegedy2016Rethinking} including 4 streams of convolutions of filter sizes $1\times1$, $3\times3$, $5\times5$ and $7\times7$. Each convolutional layer of filter larger than $1\times1$ is factorized into a sequence of layers with smaller receptive fields in order to reduce the computational cost as suggested in~\cite{Szegedy2016Rethinking}.

\subsection{Appearance convolutional autoencoder}\label{sec:convAE}
Our Conv-AE supports the detection of strange (abnormal) objects within input frames by learning common appearance templates in normal events. This sub-network consists of the encoder and the top decoder without any skip connection as shown in Figure~\ref{fig:overview}. The encoder is constructed by a sequence of blocks including triple layers: convolution, batch-normalization (BatchNorm) and leaky-ReLU activation~\cite{Maas2013Rectifier}. The first block (right after the Inception module) does not contain BatchNorm layer as suggested in~\cite{Isola2017Image} for our U-Net task in Section~\ref{sec:unet}. Instead of using pooling layer to reduce the resolution of feature maps, we apply strided convolution. Such parametric operation is expected to support the network finding an informative way to downsample the spatial resolution of feature maps as well as learning the further upsampling in decoding stage~\cite{Springenberg2014Striving}.

The decoder is also a sequence of layer blocks that increases the spatial resolution while reduces the number of feature maps after each deconvolution layer. A dropout layer (with $p_{drop}=0.3$) is attached before the ReLU activation in each block as a regularization that reduces the risk of overfitting during the training stage~\cite{Srivastava2014Dropout}.

Since the Conv-AE is to learn common appearance patterns of normal events, we consider the $l_2$ distance between the input image $I$ and its reconstruction $\hat{I}$. The model thus forces to produce an image with similar intensity for each pixel. The intensity loss is estimated as
\begin{equation}
	\mathcal{L}_{int}(I,\hat{I}) = \|I-\hat{I}\|^2_2
	\label{eq:loss_int}
\end{equation}
A drawback of using only $l_2$ loss is the blur in the output, we thus add a constraint that attempts to preserve the original gradient (\ie the sharpness) in the reconstructed image. The gradient loss is defined as the difference between absolute gradients along the two spatial dimensions as
\begin{equation}
	%\mathcal{L}_{grad}\big(I,\hat{I}\big) = \bigg{\|}\big||g_x\big(I\big)|-|g_x\big(\hat{I}\big)|\big|+\big||g_y\big(I\big)|-|g_y\big(\hat{I}\big)|\big|\bigg{\|}_1
	\mathcal{L}_{grad}(I,\hat{I}) = \sum_{d\in\{x,y\}}\bigg{\|}\big|g_d(I)\big|-\big|g_d(\hat{I})\big|\bigg{\|}_1
	\label{eq:loss_grad}
\end{equation}
where $g_d$ denotes the image gradient along the $d$-axis. The final loss function of the appearance Conv-AE is formed as a summation of the intensity and gradient losses.
\begin{equation}
	\mathcal{L}_{appe}(I,\hat{I}) = \mathcal{L}_{int}(I,\hat{I}) + \mathcal{L}_{grad}(I,\hat{I})
	\label{eq:loss_appe}
\end{equation}
This loss combination has been reported to give good performance for the task of video prediction~\cite{Mathieu2015Deep,Wen2018Future}.

\subsection{Motion prediction U-Net}\label{sec:unet}
Beside the appearance of strange object structures, unusual motions of typical objects would also be appropriate to provide an assessment of a video frame. Recall that each block in the encoder is to emphasize spatial abstractions of common objects within training frames. Our U-Net sub-network thus focuses on learning the association between such patterns and corresponding motions. The ground truth optical flow employed in this work is estimated by a pretrained FlowNet2~\cite{Eddy2017Flownet2}. Compared with related models, the optical flow outputted from FlowNet2 is not only much smoother but also preserves motion discontinuities with sharper boundaries. The motion stream is expected to associate typical motions to common appearance objects while ignoring the static background patterns.

The decoder of our U-Net has the same structure as the Conv-AE except for the skip connections. These concatenations are to combine the feature maps upsampled from a higher level of abstraction with the ones containing low-level details. The use of leaky-ReLU activation in the encoder also keeps weak responses that may be informative for the translation in the decoder.

Unlike the Conv-AE in Section~\ref{sec:convAE}, the loss between an outputted optical flow and its ground truth is measured by $l_1$ distance. There are two main reasons for this. First, the FlowNet2 model is formed as a fusion of multiple networks providing optical flows from coarse (noisy) to fine (smooth), the result might thus contain noise or even amplify noisy regions during the smoothing procedure. Second, because the selection of optical flow estimation is not limited to FlowNet2, the training ground truth obtained from other algorithms might therefore possibly have small patches of wrong and/or noisy motion measure. In order to reduce the effect of such outliers when learning the motion association, we apply $l_1$ distance loss
\begin{equation}
	\mathcal{L}_{flow}\big(F_t,\hat{F}_t\big) = \|F_t-\hat{F}_t\|_1
	\label{eq:loss_flow}
\end{equation}
where $F_t$ is the ground truth optical flow estimated from two consecutive frames $I_t$ and $I_{t+1}$, and $\hat{F}_t$ is the output of our U-Net given $I_t$. In summary, this stream attempts to predict instant motions of objects appearing in the video.%In summary, this sub-network attempts to provide common instant motions of objects appearing in the video.

\subsection{Additional motion-related objective function}
Beside the distance-based loss $\mathcal{L}_{flow}$, we also add another loss that penalizes the underlying distribution of predicted optical flow to be similar to ground truth. The generative adversarial network (GAN)~\cite{Goodfellow2014Generative} was originally introduced to allow a CNN learning an implicit distribution of patterns. The model consists of a generator that creates fake samples from noise and a discriminator that attempts to distinguish such outputs from the real patterns. Many modified GAN versions have been proposed for the task of data generation. The discriminator also plays the role of a regularization in many models. Inspired by~\cite{Mathieu2015Deep} where using a GAN loss is reported to provide better results compared with employing only distance-based ones, we apply such strategy as an additional objective function.
\begin{figure}%[t]
\begin{center}
\begin{picture}(250,100)
\put(0,-12){\includegraphics[width=\columnwidth]{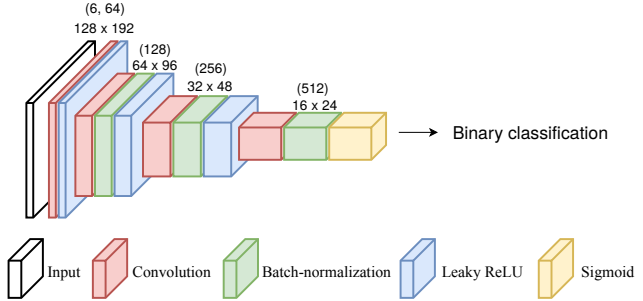}}
\end{picture}
\end{center}
\caption{The architecture of our discriminator. The input layer of shape $128\times192\times6$ is fed by the concatenation of a video frame and its optical flow (that is either ground truth or outputted from the U-Net). The output layer is sigmoid activation of 512 feature maps of spatial resolution $16\times24$. Best viewed in color.}
\label{fig:discriminator}
\end{figure}

Our generator is the entire network in Figure~\ref{fig:overview} while the discriminator conditionally performs the classification on predicted optical flow. A visualization of our discriminator architecture is shown in Figure~\ref{fig:discriminator}. Notice that the discriminator is not employed in the inference stage. Although the recent study~\cite{Wen2018Future} employed a Least Square GAN~\cite{Mao2017Least} and achieved state-of-the-art performance in detecting anomalous video frames, our model follows the strategy of typical conditional GAN (cGAN) where both the ground truth video frame and its corresponding optical flow are fed into the discriminator. There are two reasons leading to this decision. First, the cGAN theoretically avoids the problem of mode collapse in vanilla GAN since ground truth information (\ie labels, real samples) is fed into the discriminator. The model is thus expected to efficiently learn the distribution of training samples. Second, cGAN is appropriate for a CNN of image translation as demonstrated in~\cite{Isola2017Image}.

Finally, the adversarial loss is directly computed on the last layer containing activated feature maps in the discriminator. This calculation is different from~\cite{Isola2017Image,Wen2018Future} where a convolutional layer is employed to collapse previous feature channels into a 2D map. The common sense of our model and the two others is the structural penalization where the classification is performed according to image patches instead of the whole image. However, we strictly constrain patches at feature-level so that each feature map must attempt to provide a classification result. This design is inspired from the study~\cite{Long2017SCA} demonstrating that each convolutional channel attends to particular semantic patterns.

Given an input video frame $I$ and its associated optical flow $F$ obtained from FlowNet2, the proposed network in Figure~\ref{fig:overview} (the generator denoted as $\mathcal{G}$) produces a reconstructed frame $\hat{I}$ and a predicted optical flow $\hat{F}$, while the discriminator $\mathcal{D}$ estimates a probability that the optical flow associated to $I$ is the ground truth $F$. The GAN objective function consists of two loss functions:
\begin{equation}
\begin{split}
	\mathcal{L}_\mathcal{D}(I,F,\hat{F}) = & ~~~~~\frac{1}{2}\sum_{x,y,c}-\mathrm{log}\mathcal{D}(I,F)_{x,y,c} \\ &+ \frac{1}{2}\sum_{x,y,c}-\mathrm{log}[1-\mathcal{D}(I,\hat{F})_{x,y,c}]
\end{split}
	\label{eq:Dloss}
\end{equation}
\begin{equation}
\begin{split}
	\mathcal{L}_\mathcal{G}(I,\hat{I},F,\hat{F}) = & ~~~~~\lambda_{\mathcal{G}}\sum_{x,y,c}-\mathrm{log}\mathcal{D}(I,\hat{F})_{x,y,c} \\ &+ \lambda_a\mathcal{L}_{appe}(I,\hat{I}) + \lambda_f\mathcal{L}_{flow}(F,\hat{F})
\end{split}
	\label{eq:Gloss}
\end{equation}
where $x,y$ and $c$ respectively indicate the spatial position and the corresponding channel of a unit in the feature maps outputted from $\mathcal{D}$, and $\lambda$ values are the weights associated to partial losses within our proposed model. Our GAN is optimized by alternately minimizing the two GAN losses. In our experiments (see Section~\ref{sec:experiment}), we assigned 0.25 for $\lambda_{\mathcal{G}}$, 1 for $\lambda_a$ and 2 for $\lambda_f$. This GAN aims to emphasize the efficiency of motion prediction.

\subsection{Anomaly detection}\label{sec:detection}
Our model aims to provide a score of normality for each frame. In related studies, such scores are usually quantities measuring the similarity between a ground truth and the reconstructed/predicted output. There are two common scores employed in CNN approaches: $L_p$ distance and Peak Signal to Noise Ratio (PSNR). The normality of each video frame is decided by comparing its score with a threshold. It is obvious that an anomalous event occurring within a small image region may be missed due to the summation and/or average operations over all pixel positions. We hence propose another score estimation scheme considering only a small patch instead of the entire frame.

First, we define partial scores individually estimated on the two model streams sharing the same patch position as
\begin{equation}
 \begin{cases}\mathcal{S}_{I}(P) = \frac{1}{|P|} \sum_{i,j\in P} (I_{i,j}-\hat{I}_{i,j})^2 \\\mathcal{S}_{F}(P) = \frac{1}{|P|} \sum_{i,j\in P} (F_{i,j}-\hat{F}_{i,j})^2\end{cases}
	\label{eq:partialscore}
\end{equation}
where $P$ indicates an image patch and $|P|$ is its number of pixels. Our frame-level score is then computed as a weighted combination of the two partial scores as follows:
\begin{equation}
	\mathcal{S}=\log[w_F\mathcal{S}_F(\tilde{P})] +\lambda_{\mathcal{S}}\log[w_I\mathcal{S}_I(\tilde{P})]
	\label{eq:score}
\end{equation}
where $w_F$ and $w_I$ are the weights calculated according to the training data, $\lambda_{\mathcal{S}}$ is to control the contribution of partial scores to the summation, and $\tilde{P}$ is the patch providing the highest value of $\mathcal{S}_F$ in the considering frame, \ie
\begin{equation}
	\tilde{P}\leftarrow\argmax_{P \mathrm{~slides~on~frame}}\mathcal{S}_F(P)
\end{equation}
The weights $w_F$ and $w_I$ are estimated as the inverse of average scores obtained on the training data of $n$ images:
\begin{equation}
	\begin{cases}w_F=\bigg[\frac{1}{n}\sum_{i=1}^n\mathcal{S}_{F_i}(\tilde{P}_i)\bigg]^{-1}\\ w_I=\bigg[\frac{1}{n}\sum_{i=1}^n\mathcal{S}_{I_i}(\tilde{P}_i)\bigg]^{-1}\end{cases}
\end{equation}
This helps to normalize the two scores on the same scale. The size of $P$ was set to $16\times16$ in our experiments. Typically, such patches are determined by a sliding window. In realistic implementation, it can be performed using a convolutional operation with a filter of size $16\times16$. $\lambda_{\mathcal{S}}$ was empirically set to 0.2 since the model focuses on motion prediction efficiency.

Finally, we perform a normalization on frame-level scores in each evaluated video as suggested in related studies such as~\cite{Hasan2016Learning,Ravanbakhsh2017Abnormal,Wen2018Future}. Our final frame-level score is
\begin{equation}
	\mathcal{\hat{S}}_t=\frac{\mathcal{S}_t}{\max(\mathcal{S}_{1..m})}
	\label{eq:normalization}
\end{equation}
where $t$ is the frame index in a video containing $m$ frames. The score estimated from a frame of abnormal event is expected to be higher compared with the ones of normal event.

\section{Experiments}\label{sec:experiment}
\begin{figure}%[t]
\begin{center}
%\begin{picture}(250,62)
%\put(0,-7){
\includegraphics[width=\columnwidth]{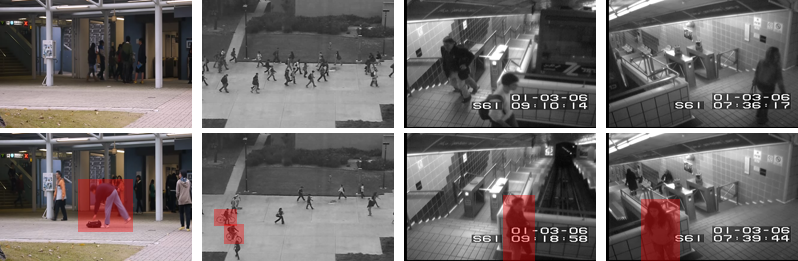}
%}
%\end{picture}
\end{center}
\caption{Examples of normal (top) and abnormal (bottom) frames in the CUHK Avenue, UCSD Ped2, Exit Gate, and Entrance Gate (from left to right) datasets. Anomalous events are highlighted including a man picking a bag, bicycle appearance, and loitering.}
\label{fig:dataset}
\end{figure}

We performed experiments on various benchmark datasets of anomaly detection including CUHK Avenue~\cite{Lu2013Abnormal}, UCSD Ped2~\cite{Li2014Anomaly}, Subway Entrance Gate and Exit Gate~\cite{Adam2008Robust}, Traffic-Belleview and Traffic-Train~\cite{Zaharescu2010Anomalous}. Their training data contain only normal events. Some examples of normal and abnormal frames in the first 4 datasets are shown in Figure~\ref{fig:dataset}. The first two datasets are provided with frame-level ground truth, we thus employ area under curve (AUC) of the receiver operating characteristic (ROC) curve measured according to frame-level scores outputted from the proposed model to indicate the performance. The next two Subway datasets are evaluated on event-level that requires some additional operations described below. The last two datasets are evaluated according to the average precision (AP) since the precision-recall (PR) curve was usually used for their assessment~\cite{Zaharescu2010Anomalous,Xu2017Detecting}. We used the FlowNet2 pretrained on FlyingThing3D~\cite{Mayer2016A} and ChairsSDHom~\cite{Eddy2017Flownet2} datasets as the ground truth optical flow estimator. The GAN was trained using Adam algorithm~\cite{Diederik2014Adam} where the initial learning rates were set to $2\times10^{-4}$ for the generator $\mathcal{G}$ and $2\times10^{-5}$ for the discriminator $\mathcal{D}$. The description, experimental results and a discussion corresponding to each evaluation are presented in the remaining of this section.

\subsection{CUHK Avenue and UCSD Ped2}

\begin{table}
\begin{center}
	\begin{tabularx}{\columnwidth}{ |l| *{2}{Y|} }%{|l|c|c|}
	\hline
	Method & Avenue & Ped2 \\
	\hline\hline
	%Social force~\cite{Mehran2009Abnormal} & - & 0.556 \\
	%MPPCA~\cite{Kim2009Observe} & - & 0.693 \\
	%MPPC + Social force~\cite{Mahadevan2010Anomaly} & - & 0.613 \\
	%MDT~\cite{Mahadevan2010Anomaly} & - & 0.829 \\
	%HOFME~\cite{Wang2012Histograms} & - & 0.875 \\
	Conv-AE~\cite{Hasan2016Learning} & 0.702 & 0.900 \\
	Discriminative learning~\cite{Giorno2016A} & 0.783 & - \\
	Hashing filters~\cite{Zhang2016Video} & - & 0.910 \\
	%Unmask conv5~\cite{Ionescu2017Unmasking} & 0.805 & 0.821 \\
	%Unmask 3D gradient~\cite{Ionescu2017Unmasking} & 0.801 & 0.813 \\
	Unmask late fusion~\cite{Ionescu2017Unmasking} & 0.806 & 0.822 \\
	%Early fusion~\cite{Xu2017Detecting} & - & 0.815 \\
	%Late fusion~\cite{Xu2017Detecting} & - & 0.873 \\
	AMDN (double fusion)~\cite{Xu2017Detecting} & - & 0.908 \\
	ConvLSTM-AE~\cite{Luo2017Remembering} & 0.770 & 0.881 \\
	DeepAppearance~\cite{Sorina2017Deep} & 0.846 & - \\
	FRCN action~\cite{Hinami2017Joint} & - & 0.922 \\
	TSC~\cite{Luo2017A} & 0.806 & 0.910 \\
	Stacked RNN~\cite{Luo2017A} & 0.817 & 0.922 \\
	AbnormalGAN~\cite{Ravanbakhsh2017Abnormal} & - & 0.935 \\
	GrowingGas~\cite{Sun2017Online} & - & 0.941 \\
	Future frame prediction~\cite{Wen2018Future} & 0.851 & 0.954 \\
	\hline
	Our proposed method & 0.869 & 0.962 \\%Avenue: 25 epochs, Ped2: 15 epochs
	\hline
	\end{tabularx}
\end{center}
\caption{Frame-level performance (AUC) of anomaly detection on the CUHK Avenue and UCSD Ped2 datasets. The methods are ordered according to the year of publication.}
\label{table:avenueped2}
\end{table}
\begin{figure}[t]
\begin{center}
\begin{picture}(250,482)
	\put(0,378){\includegraphics[width=\columnwidth]{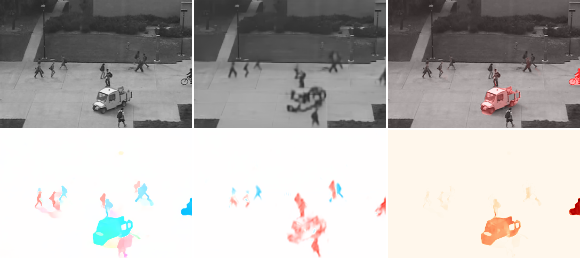}}
	\put(16,368){(a) The appearance of a truck and a bicycle. (Ped2)}
	\put(0,253){\includegraphics[width=\columnwidth]{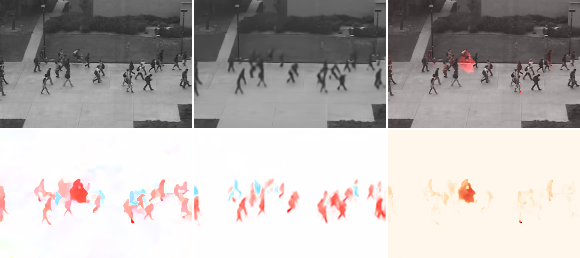}}
	\put(6,243){(b) A bicycle is running in a low contrast region. (Ped2)}
	\put(0,128){\includegraphics[width=\columnwidth]{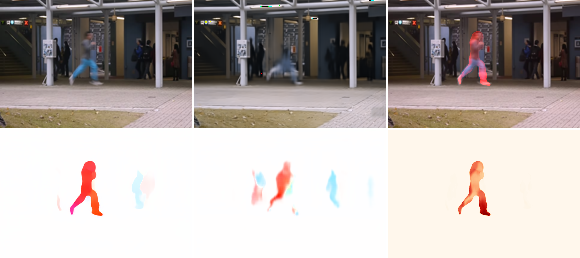}}
	\put(55,118){(c) A man is running. (Avenue)}
	\put(0,3){\includegraphics[width=\columnwidth]{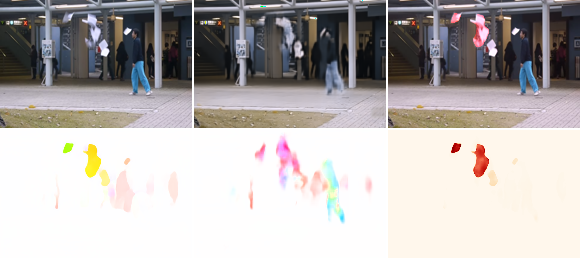}}
	\put(40,-7){(d) A man is tossing papers. (Avenue)}
\end{picture}
\end{center}
\caption{(Best viewed in color) Results on the Ped2 and Avenue datasets. Each example consists of 3 image columns that are input frame and its optical flow (left), reconstructed frame and predicted motion (middle), and the frame superimposed by the motion error map below (right). The flow field color coding is the same as~\cite{Eddy2017Flownet2}.}
\label{fig:avenue_ped2}
\end{figure}

The Avenue dataset consists of 30652 frames that are split into 16 clips for training and 21 clips for testing. This dataset was captured in a campus avenue and contains various types of anomaly such as unusual action (\eg running), wrong moving direction and abnormal object (\eg bicycle). This also provides some challenges for evaluation such as slight camera shake and the occurrence of a few outliers.

The UCSD anomaly dataset includes two subsets Ped1 and Ped2 acquired from static cameras overlooking pedestrian walkways. The anomalies are the appearance of non-pedestrian object (\eg vehicle) and strange pedestrian motion. The difference between the two subsets is the walking direction (toward and away from the camera in Ped1, parallel to the camera plane in Ped2). We select only the Ped2 dataset for two reasons. First, our optical flow estimator (FlowNet2) does not work well on very small and thin pedestrians appearing too far from the camera. Nevertheless, examples of people walking towards and away from the camera are available in the CUHK Avenue dataset allowing to evaluate performance in this situation. Second, we observed that some events were labeled as normality in the training data but were considered as anomalous in the test data (\eg people walking on grass). Therefore, the Ped2 dataset (16 training and 12 testing clips) was used in our experiments.

The frame-level assessment results in Table~\ref{table:avenueped2} show that our model outperforms all other recent methods in the task of anomaly detection. Examples of reconstructed frames and predicted optical flows obtained from the appearance and motion streams are given in Figure~\ref{fig:avenue_ped2}. Considering the first example, the truck was reconstructed as a collection of pedestrian patterns since it is a new object observed by the model. The corresponding predicted motion was thus completely different from the ground truth. The processing of the bicycle on the right image edge was also similar. The second scene shows that the model still worked well on a crowded scene with many pedestrians and an anomalous object having similar intensities with the background. In the next two Avenue frames, the model expected slower moving speed and another motion direction as observed in the training data. In addition, notice that the reconstructed man's trouser color was slightly different from the input frame while the back ground was well restored. This demonstrates that the model reasonably determined the low-significance relation between the color of a pattern and its movement.

\subsection{Subway Entrance and Exit gates}
This dataset contains videos capturing the entrance gate and exit gate of a subway station. Their lengths are respectively 96 and 43 minutes. The anomalous events in these two videos are wrong direction (\eg passenger exits through the entrance gate), no payment, loitering, irregular interaction (\eg a person walks awkwardly to avoid another) and miscellaneous (\eg sudden changing of walking speed).

We performed the evaluation according to the ground truth of events with the training and test sets provided in~\cite{Kim2009Observe}, in which the normal events in the first 15 minutes of the Entrance Gate video and 5 minutes of the Exit Gate were used in training stage. Notice that the experiments were performed individually for the two videos.

Since the dataset does not provide the frame-level ground truth, we employ the assessment scheme in~\cite{Hasan2016Learning} to determine anomalous events in the experiments. In detail, the persistence algorithm~\cite{persistence1d} is applied on the sequence of scores to locate local maxima, in which each maximum point indicates an anomalous event. In order to reduce the effect of possible noisy detected extrema, nearby events are combined to provide only an anomalous one.% persistence threshold: 19 and 11.75

\begin{table}
\begin{center}
\begin{tabularx}{\columnwidth}{ |l| *{4}{Y|} }
\hline
\multicolumn{1}{|l|}{\multirow{2}{*}{Method}} & \multicolumn{2}{c|}{Entrance (66)} & \multicolumn{2}{c|}{Exit (19)} \\ \cline{2-5} 
\multicolumn{1}{|l|}{} & \multicolumn{1}{c|}{TP} & \multicolumn{1}{c|}{FA} & \multicolumn{1}{c|}{TP} & \multicolumn{1}{c|}{FA} \\
\hline\hline
Subspace~\cite{Elhamifar2009Sparse} & 46 & 7 & 14 & 4 \\
MPPCA~\cite{Kim2009Observe} & 57 & 6 & 19 & 3 \\
DSC~\cite{Zhao2011Online} & 60 & 5 & 19 & 2 \\
%STC~\cite{Roshtkhari2013Online} & 60 & 4 & 19 & 2 \\
Sparse dict.~\cite{Lu2013Abnormal} & 57 & 4 & 19 & 2 \\
Conv-AE~\cite{Hasan2016Learning} & 61 & 15 & 17 & 5 \\
IT-AE~\cite{Hasan2016Learning} & 55 & 17 & 17 & 9 \\
Hashing filters~\cite{Zhang2016Video} & 61 & 4 & 19 & 2 \\
Early fusion~\cite{Xu2017Detecting} & 56 & 8 & 15 & 4 \\
Late fusion~\cite{Xu2017Detecting} & 58 & 6 & 17 & 2 \\
AMDN~\cite{Xu2017Detecting} & 61 & 4 & 19 & 1 \\
\hline
Our method & 61 & 18 & 17 & 5 \\
\hline
\end{tabularx}
\end{center}
\caption{Our results of anomaly detection on the Subway datasets. In the ground truth, the numbers of abnormal events in the Entrance and Exit are respectively 66 and 19. The term TP indicates the number of true positive detections while FA is the counting of false alarms. The methods are listed in temporal order.}%The methods are ordered by to the year of publication.}
\label{table:subway}
\end{table}
\begin{figure*}[t]
\begin{center}
\begin{picture}(520,130)
	% false alarm
	\put(250,2){\includegraphics[scale=0.25]{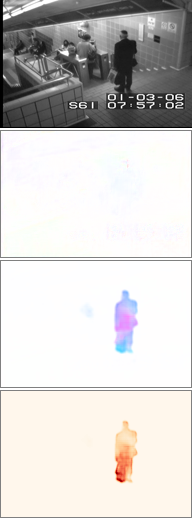}}
	\put(300,2){\includegraphics[scale=0.25]{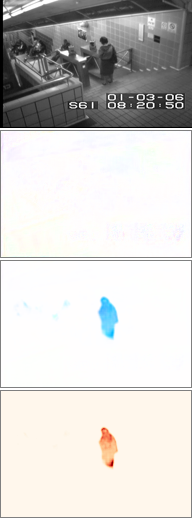}}
	\put(350,2){\includegraphics[scale=0.25]{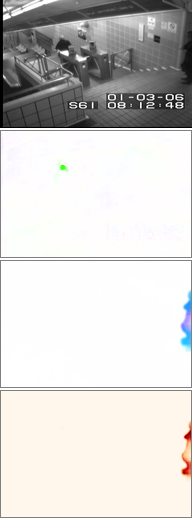}}
	\put(400,2){\includegraphics[scale=0.25]{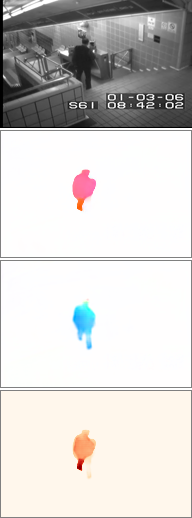}}
	\put(450,2){\includegraphics[scale=0.25]{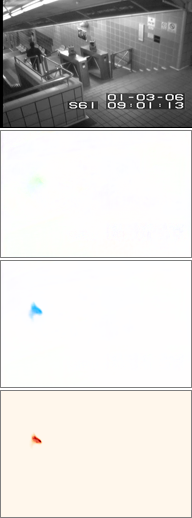}}
	\put(270,-8){(f)}\put(320,-8){(g)}\put(370,-8){(h)}\put(421,-8){(i)}\put(470,-8){(j)}
	% missed detection
	\put(0,2){\includegraphics[scale=0.25]{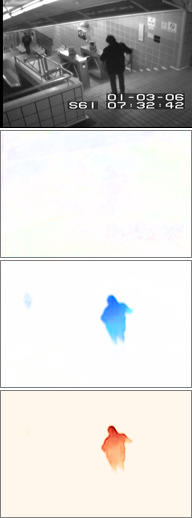}}
	\put(50,2){\includegraphics[scale=0.25]{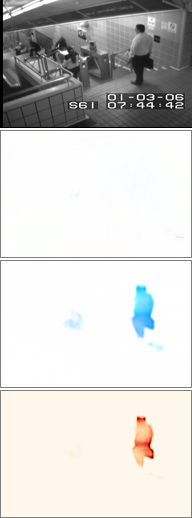}}
	\put(100,2){\includegraphics[scale=0.25]{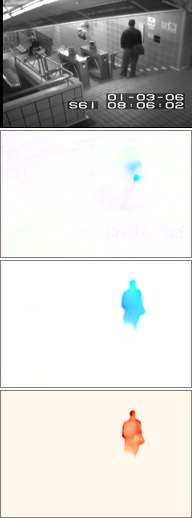}}
	\put(150,2){\includegraphics[scale=0.25]{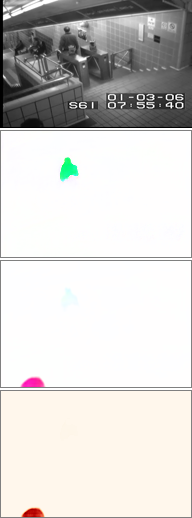}}
	\put(200,2){\includegraphics[scale=0.25]{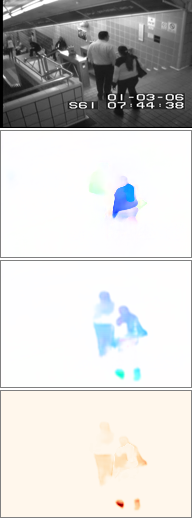}}
	\put(21,-8){(a)}\put(71,-8){(b)}\put(120,-8){(c)}\put(170,-8){(d)}\put(220,-8){(e)}
\end{picture}
\end{center}
\caption{Examples of missed detections (a)-(e) and false alarms (f)-(j) in our experiments on the Entrance dataset. Each example consists of 4 images that are (from top to bottom) the input frame, ground truth optical flow, predicted motion and the corresponding motion error map. The missed detections are: (a)-(c) movement stopping, (d) loitering, and (e) loitering (man) and movement stopping (woman). The false alarms are: (f)-(g) movement stopping, (h) loitering, (i) changing gate, and (j) passenger going near the railway. Best viewed in color.}
\label{fig:subway}
\end{figure*}
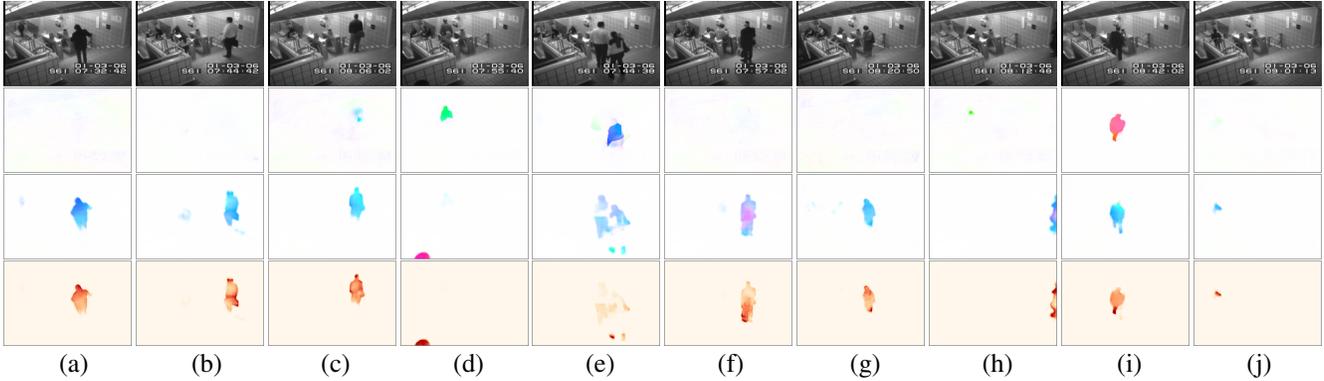
Our event-based assessment results are presented in Table~\ref{table:subway}. It shows that our model detected most anomalous events but also generated more false alarm than other recent studies. By taking a closer look at these false alarms, we determined that some events denoted as normal in the test set can be considered as anomaly under other circumstances. A visualization of some false alarms and missed anomaly detections in the Entrance dataset is given in Figure~\ref{fig:subway}.

Figure~\ref{fig:subway} shows that the normality decision of movement stopping and loitering was unstable since the cases (a)-(e) were missed while (f)-(h) were wrongly detected. There are two possible reasons: (1) the use of maximum localization as in~\cite{Hasan2016Learning} is not ideal when the anomaly score smoothly and/or slowly changes, and (2) the training set (according to~\cite{Kim2009Observe}) contains loitering event [caused by the man in (b) and (e)]. The ambiguity in ground truth annotation is also shown in the event (h) where a loitering man appeared on the right side but was not labeled as anomaly. In the event (i), the model predicted that the man would go through the left gate but he suddenly changed to the right one (the color indicates the motion direction). Since this action does not occur in the training data, the model determined it as an anomalous event. Regarding the last example (j), the motion stream expected the passenger to go to the train because most people at this location move to the left side in the training data. In other words, the model may \textit{forget} training patterns moving to the right side. In this case, using sparse coding approaches~\cite{Kim2009Observe,Zhao2011Online,Lu2013Abnormal} can be appropriate since the effect of the frequency of training patterns is reduced.

\subsection{Traffic-Belleview and Traffic-Train}
The Traffic-Belleview dataset was acquired by a surveillance camera looking at the traffic on a road intersection from a high viewpoint. In the training data (300 frames), vehicles only run on the main street. The appearance and movement of vehicles from/to left or right roads is defined as anomaly in the test set containing a total of 2618 frames. The video is gray-scale and has a low quality.

Unlike the previous benchmark datasets, the Traffic-Train can be considered as the most challenging dataset since the lighting conditions vary drastically together with camera jitter. The camera was mounted in a train and people movement is defined as anomaly. The training and test sets consist of 800 and 4160 frames, respectively.

\begin{figure}[t]
\begin{center}
\begin{picture}(250,303)
	\put(0,252){\includegraphics[width=\columnwidth]{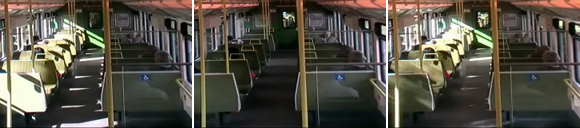}}
	\put(10,242){(a) The change of lighting in the Traffic-Train dataset.}
	\put(0,127){\includegraphics[width=\columnwidth]{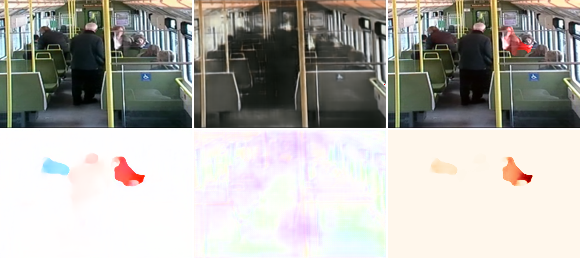}}
	\put(28.5,117){(b) Passengers moving in the stopping train.}
	\put(0,2){\includegraphics[width=\columnwidth]{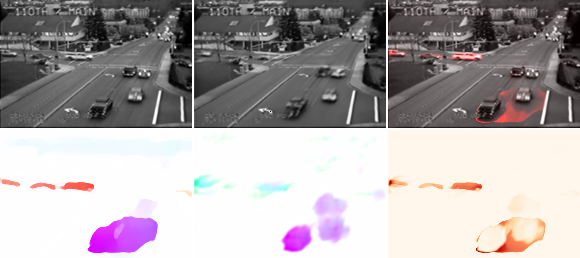}}
	\put(54,-8){(c) Cars turning to the left way.}
\end{picture}
\end{center}
\caption{(Best viewed in color) Some testing results on the two traffic datasets. Each example consists of 6 images as in Figure~\ref{fig:avenue_ped2}.}
\label{fig:belleview_train}
\end{figure}

Our average precision of frame-level assessment is presented in Table~\ref{table:belleview_train}. Figure~\ref{fig:belleview_train} shows examples of problems that the model encountered when dealing with the traffic datasets as well as illustrates the change of lighting conditions in the Train dataset. In Figure~\ref{fig:belleview_train}(b), the predicted motion was very noisy and the passenger at the frame center was missed in the error map. The effect of optical flow estimator is illustrated in Figure~\ref{fig:belleview_train}(c) where two cars were combined to be a big blob. This bad estimation significantly affected the error map though the three cars running on other way were correctly determined. The results may thus be improved by choosing another optical flow estimator or tuning the pretrained FlowNet2 by a more appropriate dataset.
\begin{table}
\begin{center}
	\begin{tabularx}{\columnwidth}{ |l| *{2}{Y|} }%{|l|c|c|}
	\hline
	Method & Belleview & Train \\
	\hline\hline
	GANomaly~\cite{Samet2018GANomaly} & 0.735 & 0.194 \\
	AEs + local feature~\cite{Narasimhan2018Dynamic} & 0.748 & 0.171 \\
	AEs + global feature~\cite{Narasimhan2018Dynamic} & 0.776 & 0.216 \\
	ALOCC $\mathcal{D}(X)$~\cite{Sabokrou2018Adversarially} & 0.734 & 0.182 \\
	ALOCC $\mathcal{D}(\mathcal{R}(X))$~\cite{Sabokrou2018Adversarially} & 0.805 & 0.237 \\
	\hline
	Our proposed method & 0.751 & 0.490 \\%Belleview: 120 epochs, Train: 25 epochs
	SSIM on appearance stream & 0.830 & 0.798 \\%Belleview: 120 epochs, Train: 25 epochs
	\hline
	\end{tabularx}
\end{center}
\caption{The average precision of frame-level anomaly detection on the Traffic-Belleview and Traffic-Train datasets.}
\label{table:belleview_train}
\end{table}

As an attempt to reduce the effect of such factors, we estimated another frame-level score without the support of motion as in section~\ref{sec:detection}. Concretely, we used the Structural Similarity Index (SSIM)~\cite{Wang2004Image} to compute the similarity between an input frame and its reconstruction provided by the appearance stream. Compared with other common measures such as MSE or PSNR, SSIM can work well on jitter images where pixel by pixel comparison is not appropriate. Table~\ref{table:belleview_train} shows that this modification improved the anomaly detection results, especially with the Train dataset.

Further details including ROC and PR curves, visualization of some feature maps and evaluation results of each single stream are provided in the supplementary materials.

\section{Conclusion}\label{sec:conclusion}
This paper presents an anomaly detection approach that exploits the correspondence between pattern appearances and their motions. The model is designed as a combination of two streams. The first one attempts to reconstruct the appearance according to its auto-encoder architecture while the second stream uses a U-Net structure to predict the instant motion given an input video frame. By sharing the same encoder, the model is forced to learn the correspondence. A patch-based scheme of anomaly score estimation is proposed to reduce the effect of noise in model outputs. Experiments on 6 benchmark datasets demonstrated the potential of our method. Detailed discussions are also presented to provide improvement suggestions for further works.

{\small
\bibliographystyle{ieee_fullname}
\bibliography{egbib}
}

\end{document}

%%%%%%%%% TITLE
\title{Anomaly Detection in Video Sequence with Appearance-Motion Correspondence\\----- Supplementary Material -----}

\author{Trong-Nguyen Nguyen, Jean Meunier\\
DIRO, University of Montreal\\
{\tt\small \{nguyetn, meunier\}@iro.umontreal.ca}
}

\maketitle
\ificcvfinal\thispagestyle{empty}\fi

%%%%%%%%% ABSTRACT
\begin{abstract}
This supplementary material provides these contents:
\begin{itemize}
	\item ROC curves of our frame-level scores on the CUHK Avenue and UCSD Ped2 datasets, and Precision-Recall (PR) curves on the traffic datasets.
	\item Experimental results of using either appearance reconstruction stream or motion prediction stream for score estimation.
	\item Impact of integrating motion stream and patch-based score estimation.
	\item Visualization of some feature maps in different blocks obtained in our experiments.
	\item Reconstructed frames and predicted motions after some training epochs.
\end{itemize}
\end{abstract}
%%%%%%%%% BODY TEXT
\section{Flow field color coding}
\begin{figure}[H]
\begin{center}
\begin{picture}(250,117)
	\put(60,0){\includegraphics[scale=0.95]{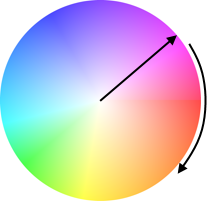}}
	\put(112,58){\rotatebox{41}{\minibox{\hspace{0.5cm}\small{pixel}\\\small{displacement}}}}
	\put(181,40){\rotatebox{90}{direction}}
\end{picture}
\end{center}
\caption{The color coding used for visualizing our optical flow in the main paper.}
\label{fig:color_coding}
\end{figure}

Figure~\ref{fig:color_coding} shows the color coding used in visualization of our optical flow in the main paper. This color coding is similar to~\cite{Eddy2017Flownet2} where the color indicates motion direction and the saturation corresponds to the pixel displacement.

\section{Evaluation curves on 4 datasets}
\begin{figure}[t]
\begin{center}
\begin{picture}(250,244)
\put(0,127){\includegraphics[scale=0.49]{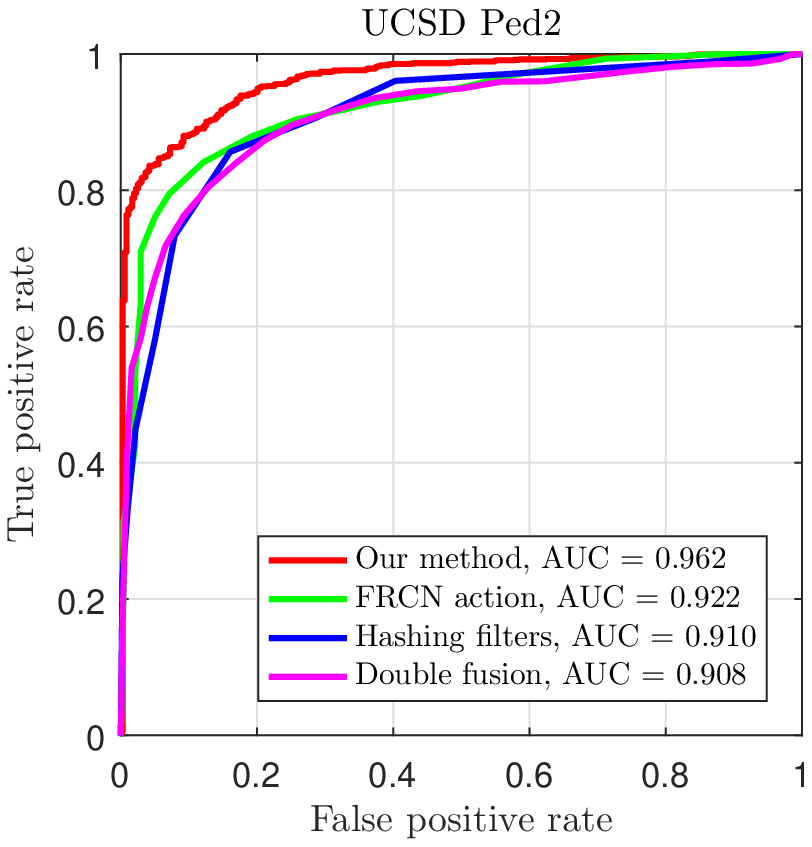}}
\put(125,127){\includegraphics[scale=0.49]{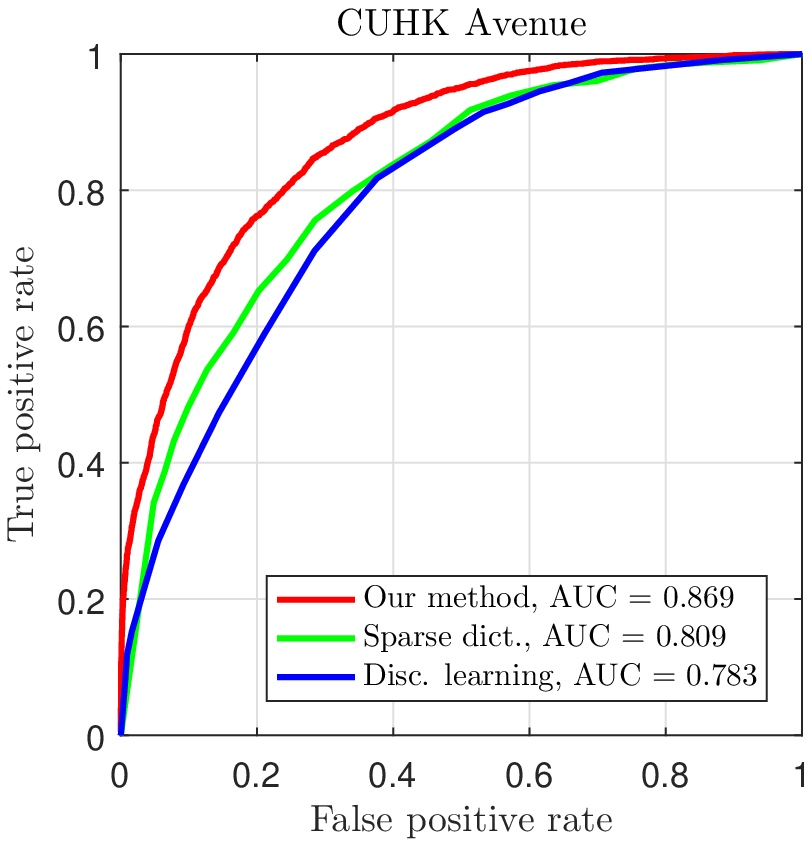}}
\put(0,0){\includegraphics[scale=0.49]{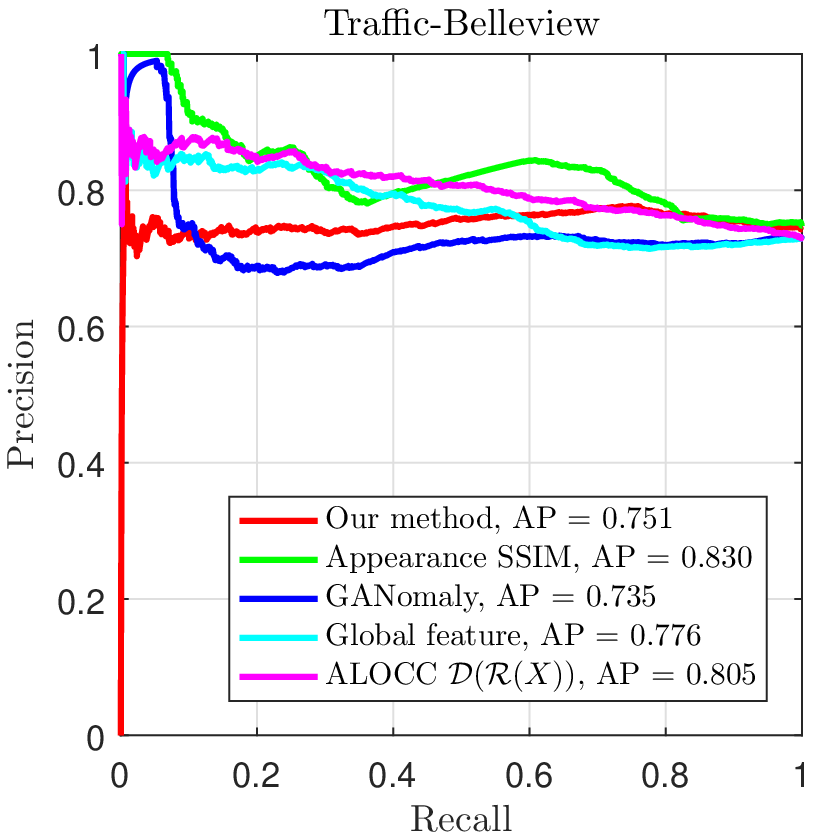}}
\put(125,0){\includegraphics[scale=0.49]{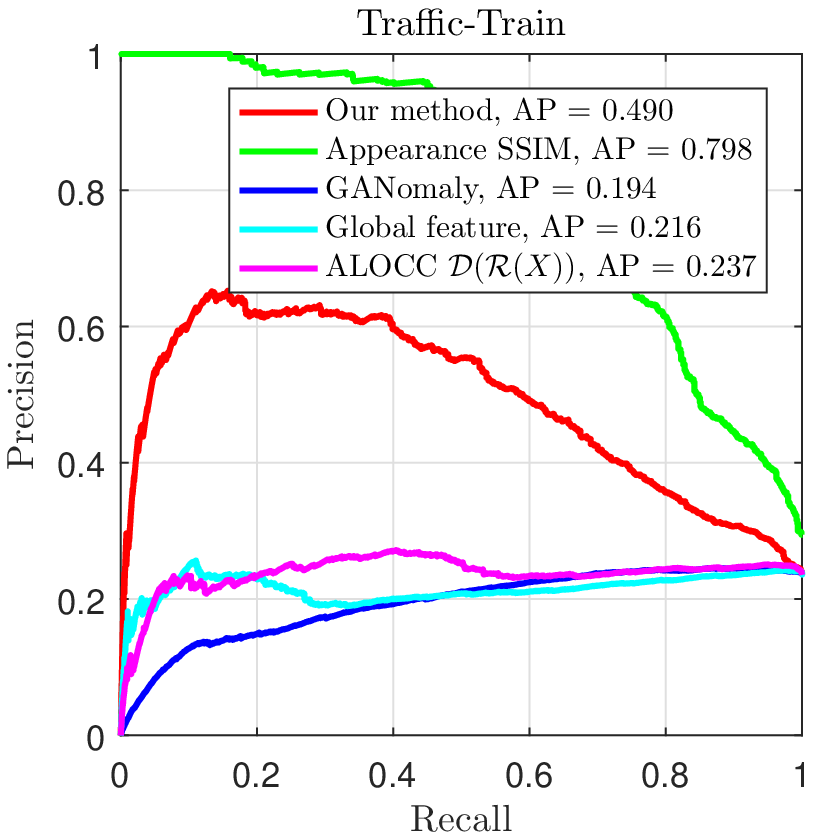}}
\end{picture}
\end{center}
\caption{Top: ROC curves on the Ped2 and Avenue datasets. Bottom: PR curves on the Belleview and Train datasets. The corresponding Area Under Curve (AUC) and Average Precision (AP) are also provided. Best viewed in color.}
\label{fig:graphs}
\end{figure}

Figure~\ref{fig:graphs} displays ROC and PR curves of our frame-level scores obtained in the experiments. Some state-of-the-art methods are also added into the figure to provide a visual comparison. These methods consist of FRCN action~\cite{Hinami2017Joint}, hashing filters~\cite{Zhang2016Video}, AMDN double fusion~\cite{Xu2017Detecting}, sparse dictionary~\cite{Lu2013Abnormal}, discriminative learning~\cite{Giorno2016A}, GANomaly~\cite{Samet2018GANomaly}, auto-encoder with global features~\cite{Narasimhan2018Dynamic} and ALOCC~\cite{Sabokrou2018Adversarially}. The ROC curves of the first 5 mentioned studies are provided in their original papers.

\section{Experimental results on single streams}
As indicated in the main paper, our frame-level score is estimated as a weighted combination of two partial scores
\begin{equation}
	\mathcal{S}=\log[w_F\mathcal{S}_F(\tilde{P})] +\lambda_{\mathcal{S}}\log[w_I\mathcal{S}_I(\tilde{P})]
	\label{eq:score}
\end{equation}
where $\mathcal{S}_F(\tilde{P})$ and $\mathcal{S}_I(\tilde{P})$ are respectively partial scores calculated from the motion and appearance streams, $w_F$ and $w_I$ are corresponding weights computed from the training data, $\lambda_{\mathcal{S}}$ is a hyperparameter controlling the contribution of partial scores to the summation, and $\tilde{P}$ is the patch providing the highest value of $\mathcal{S}_F$ in the considering frame.

\begin{figure}[t]
\begin{center}
\begin{picture}(250,373)
\put(0,254){\includegraphics[scale=0.43]{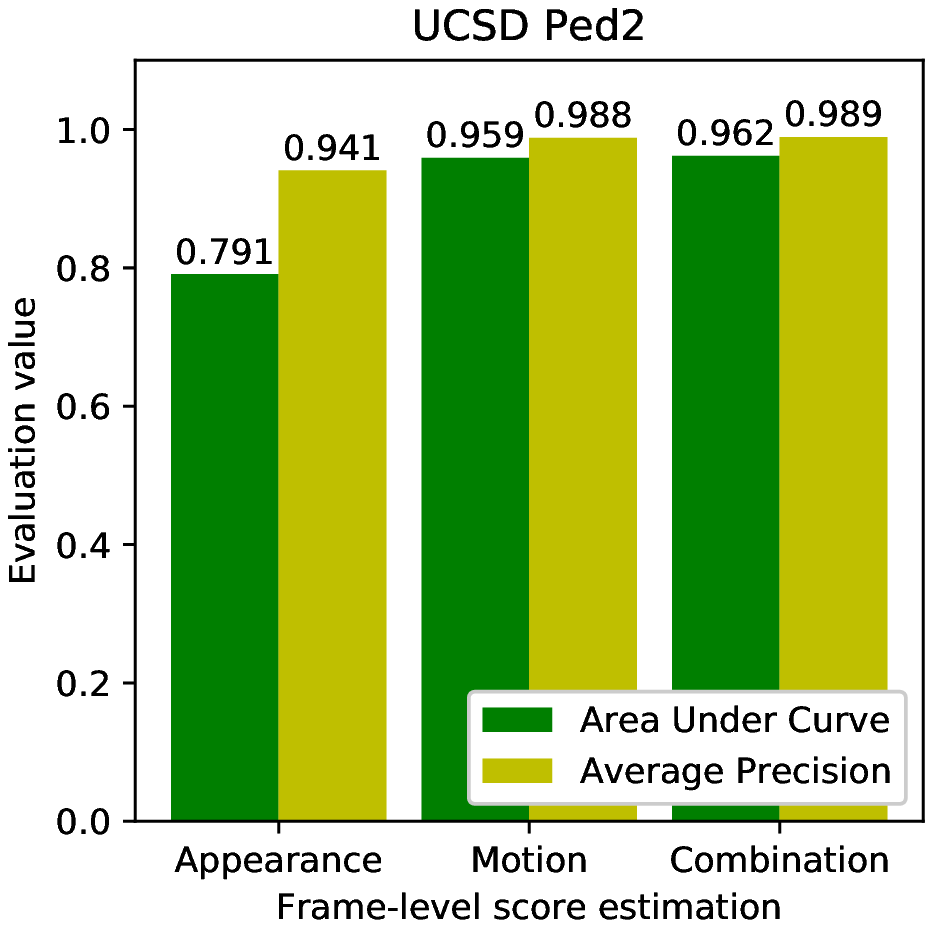}}
\put(120,254){\includegraphics[scale=0.43]{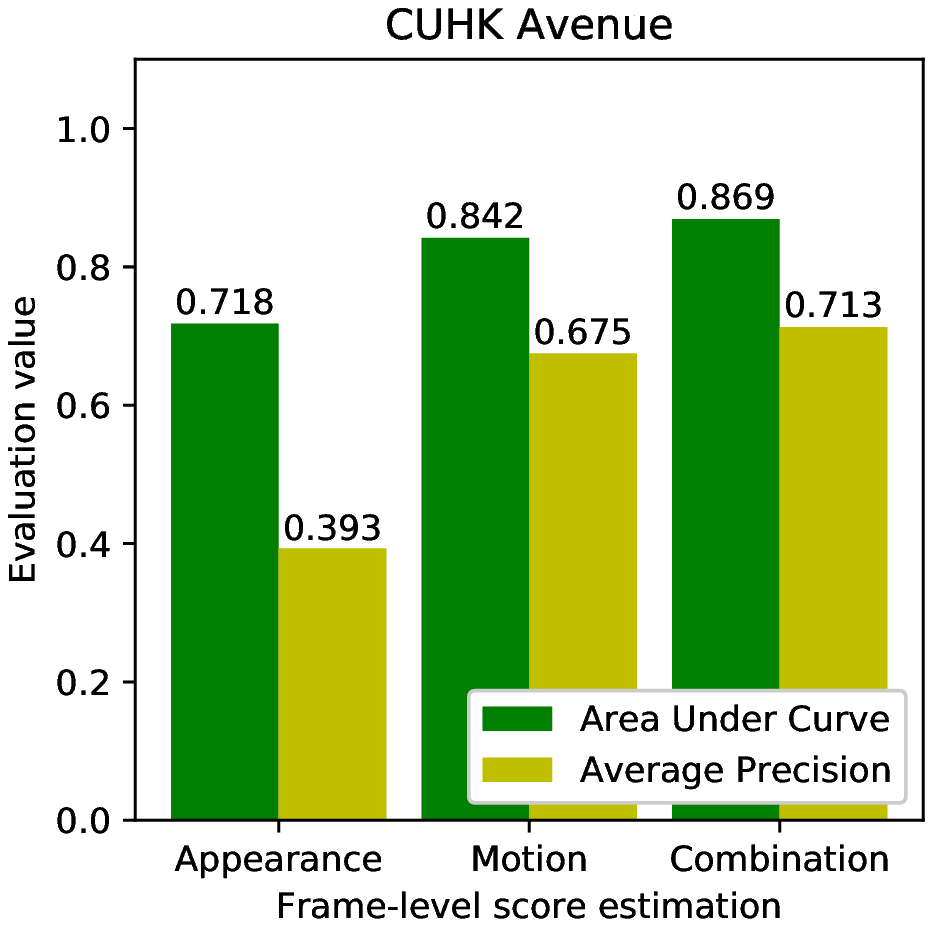}}
\put(0,127){\includegraphics[scale=0.43]{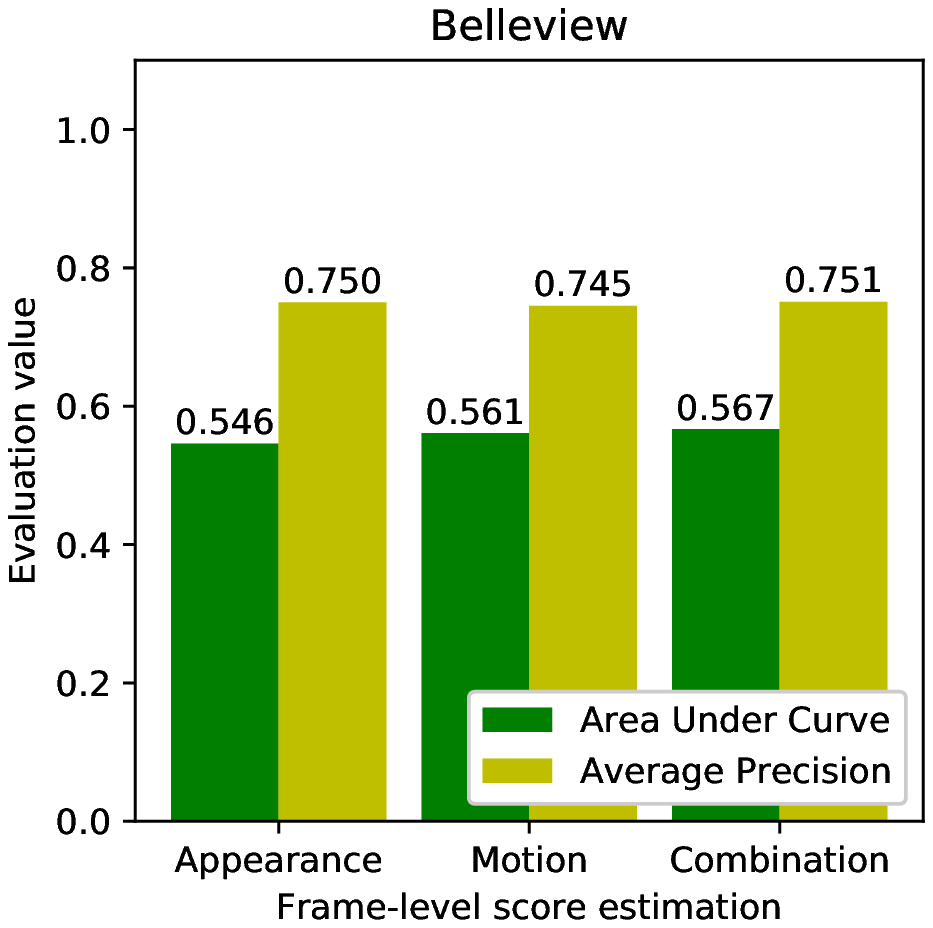}}
\put(120,127){\includegraphics[scale=0.43]{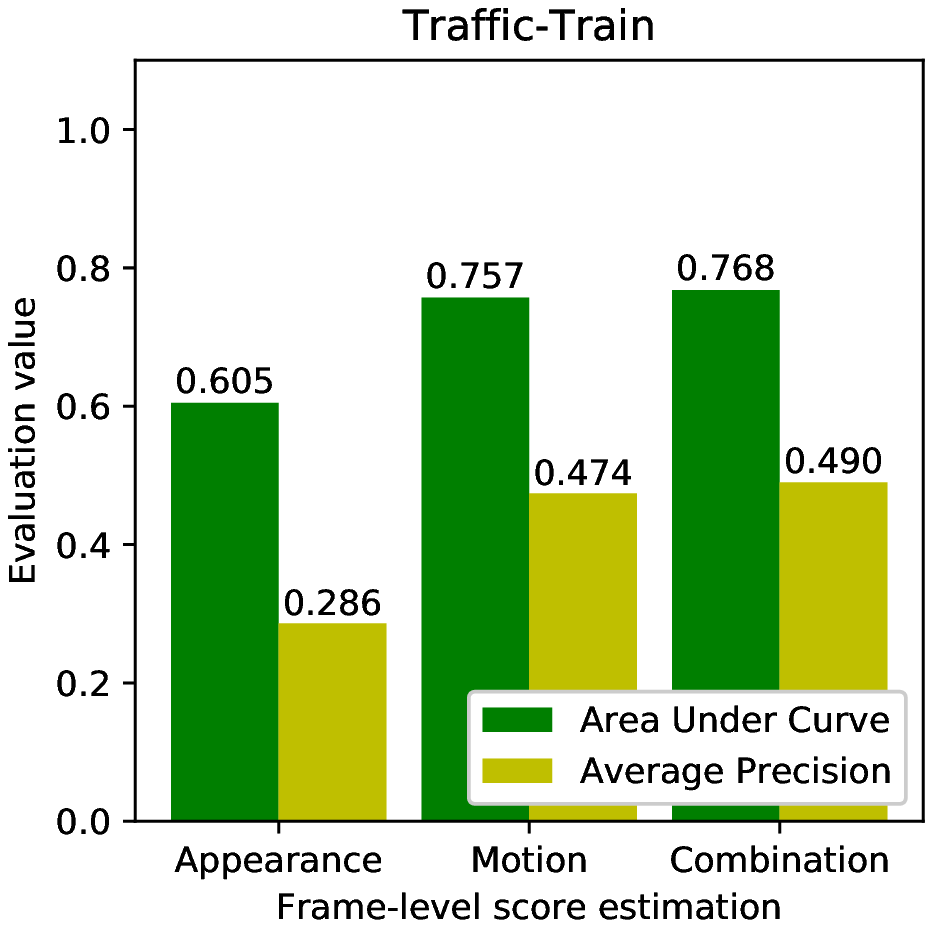}}
\put(0,0){\includegraphics[scale=0.43]{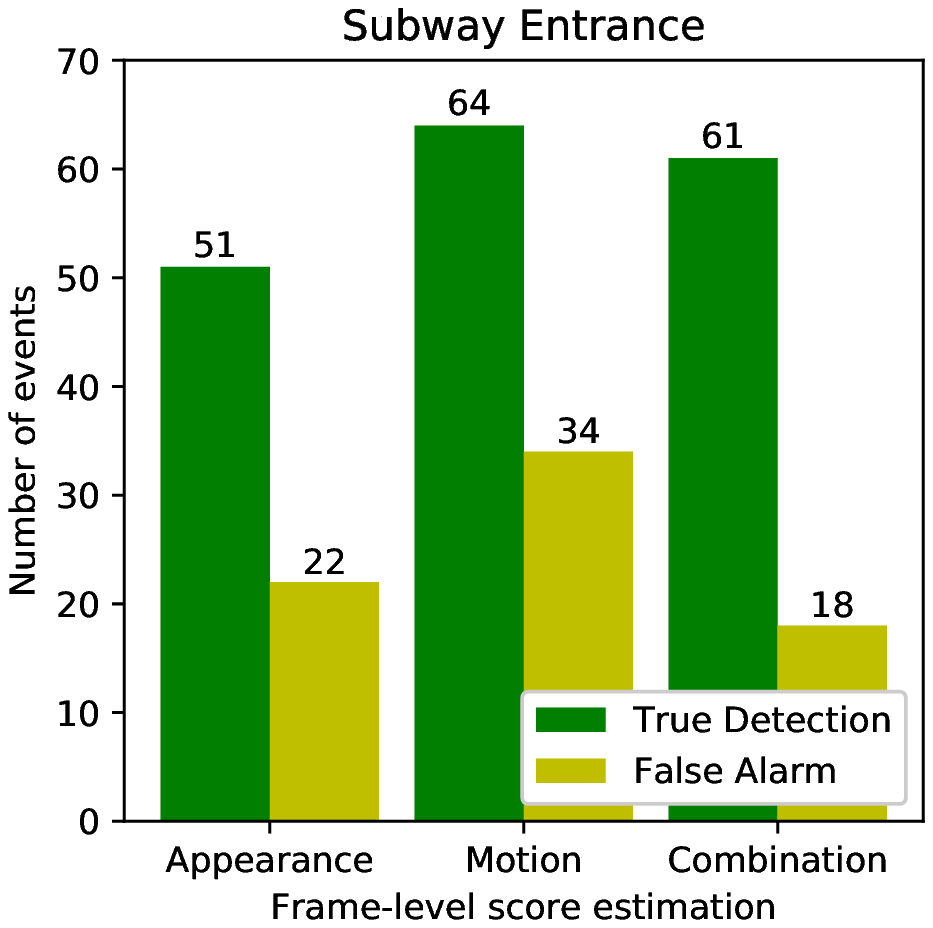}}
\put(120,0){\includegraphics[scale=0.43]{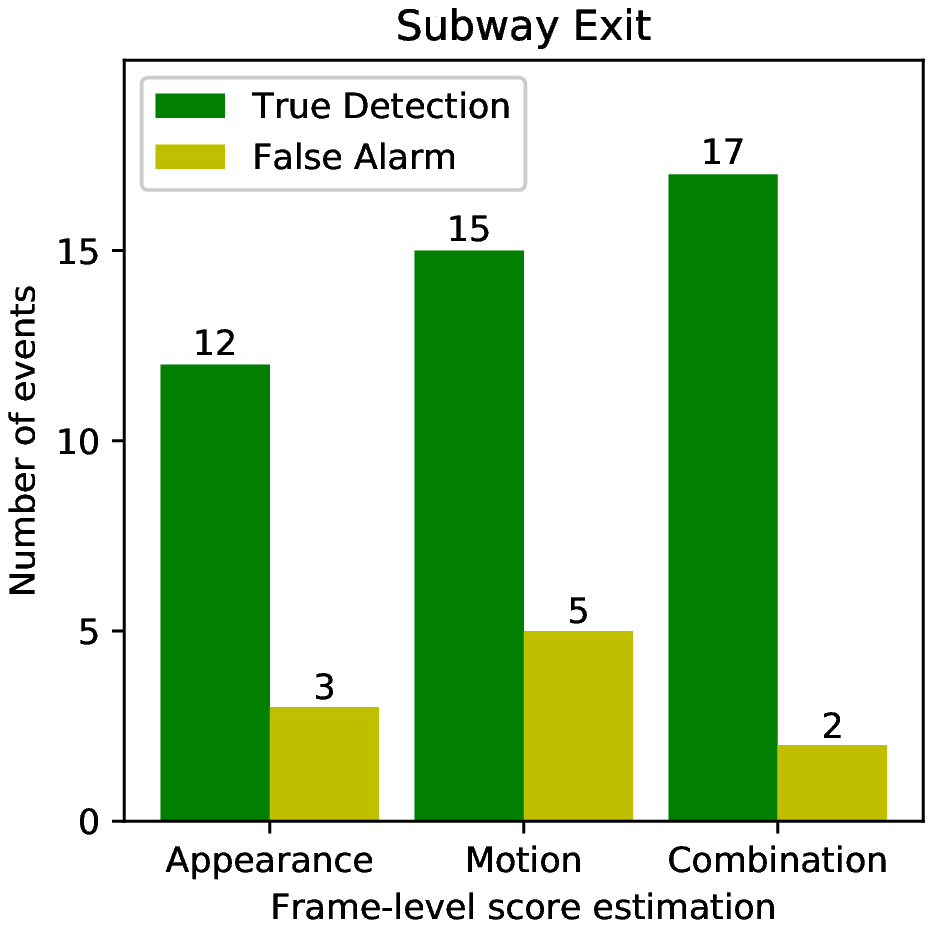}}
\end{picture}
\end{center}
\caption{Evaluation results of our model using only the appearance reconstruction (Conv-AE), the motion prediction (U-Net) and their combination. The frame-level AUROC and Average Precision scores are provided for the Ped2, Avenue, Belleview and Train datasets. The numbers of true positive detections (\ie true positive) and false alarms are presented for the Entrance and Exit datasets.}
\label{fig:single_eval}
\end{figure}

In this section, we present the evaluation results in the cases of using only one of the two partial scores as the frame-level score indicator (see Figure~\ref{fig:single_eval}). Both AUC and average precision (AP) measures are also provided for a convenient comparison with other studies. Note that the AUC and AP values are not comparable though there is a connection between ROC and PR spaces, and they are both affected by the balance of the two classes in each dataset~\cite{Davis2006The}.

Figure~\ref{fig:single_eval} shows that the combination of the two partial scores improved the detection ability since its AUC and AP increased compared with individual measures. For the Subway datasets, this combination reduced the risk of false detection, but the number of detected anomalous events was also slightly decreased (Subway Entrance).

\section{Impact of motion stream and patch-based score estimation for anomaly detection}
Table~\ref{table:impact} shows the experimental results obtained on the 6 benchmark datasets using patch-based normality assessment and SSIM on appearance stream. We also remove the motion stream and the motion-oriented discriminator (Sections 3.3 \& 3.4 in main paper) for the assessment of motion impact.
\begin{table}[h]
%\caption{Evaluation results.}
\footnotesize
\begin{threeparttable}
\begin{tabularx}{\columnwidth}{ |l| *{6}{Y|} }%{|l|c|c|}
\hline
& Avenue$^\dag$ & Ped2$^\dag$ & Entran. & Exit & Belle.$^\ddag$ & Train$^\ddag$ \\ \hline
\multicolumn{7}{|c|}{\textbf{Proposed architecture with motion stream}} \\\hline
Patch & 0.869  & 0.962 & 61/18    & 17/5 & 0.751 & 0.490\\
SSIM        & 0.694 & 0.799 & 51/14 & 15/4 & 0.830 & 0.798\\ \hline%\hline
\multicolumn{7}{|c|}{\textbf{Architecture without motion stream}} \\\hline
Patch & 0.702 & 0.773 & 58/16 & 14/7 & 0.838 & 0.380\\
SSIM  & 0.694 & 0.761 & 48/12 & 14/5 & 0.832 & 0.808\\ \hline
\end{tabularx}
\begin{tablenotes}
  \item \footnotesize{Note: True Positive / False Alarm for Entrance, Exit;~$^\dag$AUROC;~$^\ddag$AP.}
\end{tablenotes}
\end{threeparttable}
\caption{Experimental results using patch-based normality assessment and SSIM on appearance stream.}
\label{table:impact}
\end{table}
SSIM was suggested due to the errors in optical flow measurement (camera jitter in Traffic-Train and low-quality frames in Belleview). Without motion stream, the model becomes a reconstruction auto-encoder of \textit{single} frame, and the results on the first 5 datasets still demonstrate the efficiency of the proposed patch-based normality score. Using motion significantly improved results of the first 4 datasets while SSIM on appearance stream was just slightly reduced for the others (\ie 0.830 vs. 0.832 for Belleview, and 0.798 vs. 0.808 for Traffic-Train).

\section{Feature maps}
A visualization of some feature maps given an input frame for each dataset is shown in Figure~\ref{fig:feature_maps}. Each example is represented by 4 rows of images. We illustrate two feature maps (grouped in a red bounding box) for each layer block, except for the Inception module where 4 feature maps are shown for the $1\times1$, $3\times3$, $5\times5$ and $7\times7$ convolutional filters. The first two rows include the input frame, activation maps resulting from the Inception module and subsequent blocks of the shared encoder. The third and fourth rows respectively consist of feature maps in the decoder of motion and appearance streams. The value of units in each map was normalized to provide a good visualization.

Figure~\ref{fig:feature_maps} shows that our motion stream attempts to emphasize the image edges to provide a smooth optical flow (because FlowNet2~\cite{Eddy2017Flownet2} was used as the ground truth motion estimator) while the other one tends to reconstruct appearance textures. By observing all feature maps provided by the Inception module, we found that $7\times7$ convolutional filters extracted informative details only on the CUHK Avenue, Subway Entrance and Traffic-Belleview datasets (best viewed when the feature map is enlarged). It demonstrated the reasonable use of Inception module right after the input layer to let the network automatically decides its appropriate low-level filter sizes.
%\begin{figure}[t]
%\begin{center}
%{(a) UCSD Ped2}
%\includegraphics[width=\columnwidth]{figures/Ped2_2.png}
%{(b) CUHK Avenue}
%\includegraphics[width=\columnwidth]{figures/Avenue_2.png}
%{(c) Subway Entrance}
%\includegraphics[width=\columnwidth]{figures/Entrance_2.png}
%{(d) Subway Exit}
%\includegraphics[width=\columnwidth]{figures/Exit_2.png}
%{(e) Traffic-Belleview}
%\includegraphics[width=\columnwidth]{figures/Belleview_2.png}
%{(f) Traffic-Train}
%\includegraphics[width=\columnwidth]{figures/Train_2.png}
%\end{center}
%\caption{Visualization of some activation maps given an input frame for each dataset. Best viewed in color.}
%\label{fig:feature_maps}
%\end{figure}
\begin{figure*}[b]
\begin{subfigure}[t]{\textwidth}
\includegraphics[width=\textwidth]{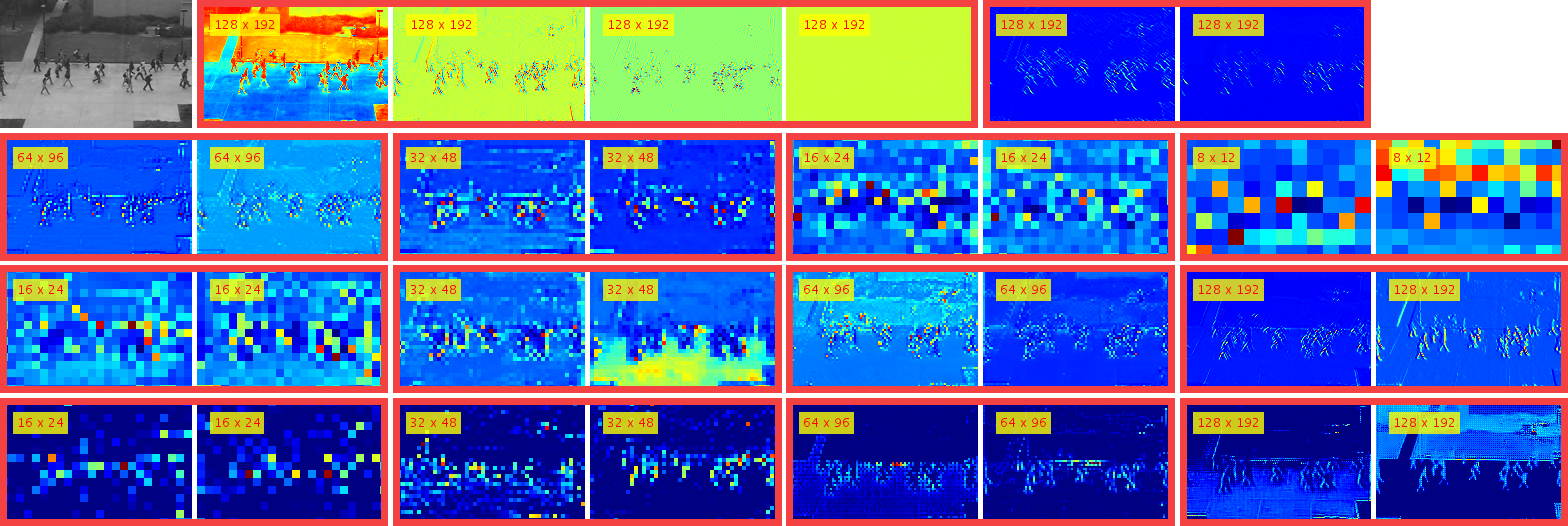}
\caption{UCSD Ped2}
\end{subfigure}
\begin{subfigure}[t]{\textwidth}
\vspace{0.7cm}
\includegraphics[width=\textwidth]{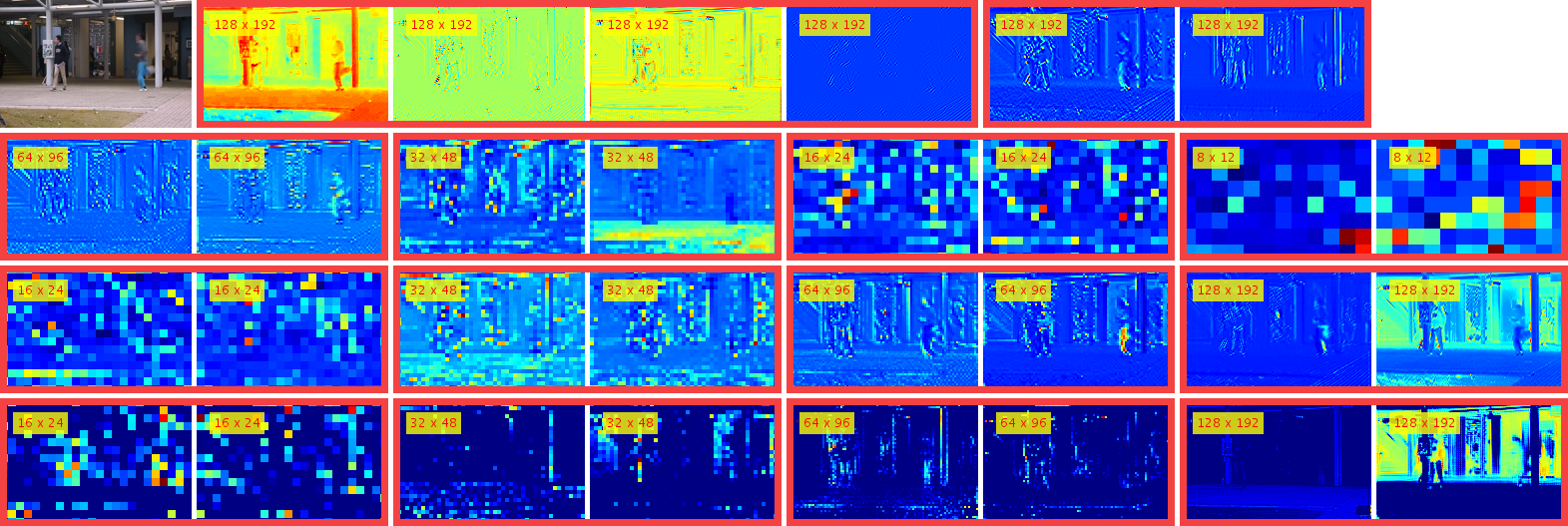}
\caption{CUHK Avenue}
\end{subfigure}
\begin{subfigure}[t]{\textwidth}
\vspace{0.7cm}
\includegraphics[width=\textwidth]{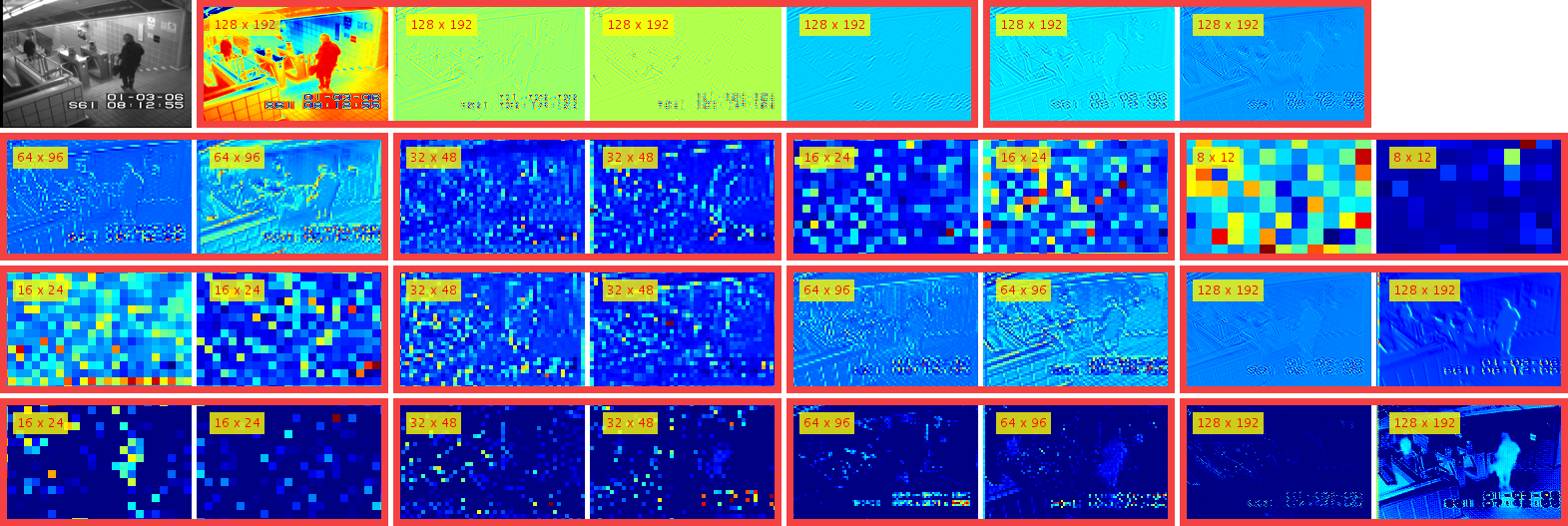}
\caption{Subway Entrance}
\end{subfigure}
\end{figure*}
\begin{figure*}\ContinuedFloat
\begin{subfigure}[t]{\textwidth}
\includegraphics[width=\textwidth]{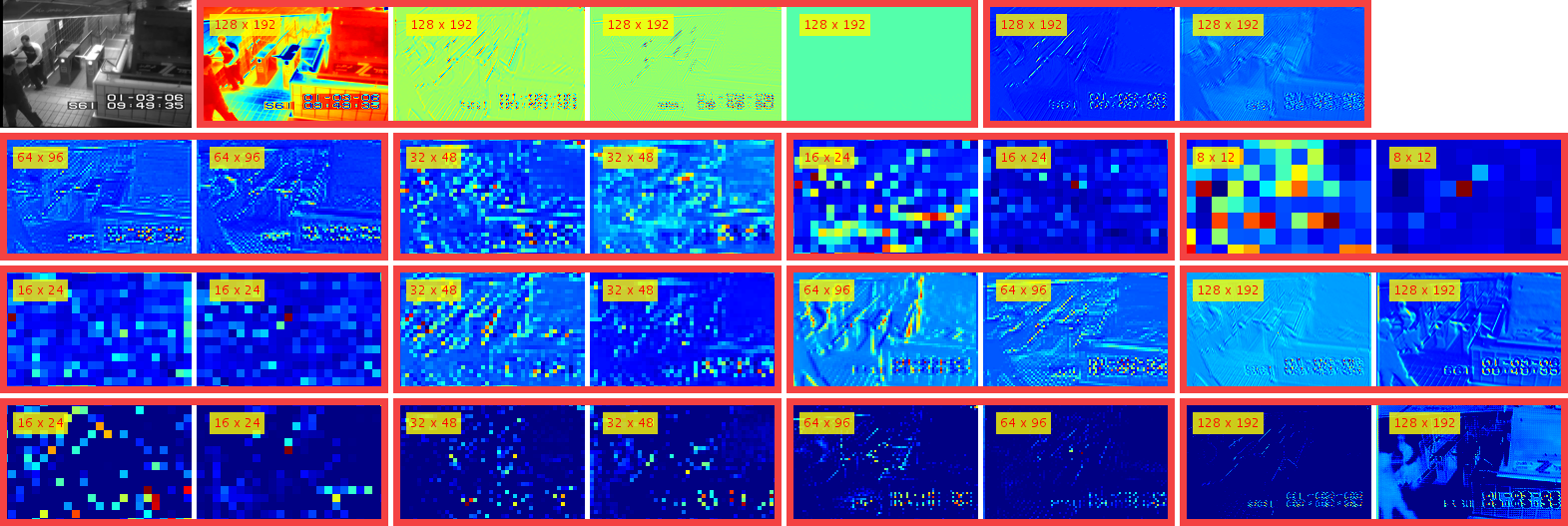}
\caption{Subway Exit}
\end{subfigure}
\begin{subfigure}[t]{\textwidth}
\vspace{0.7cm}
\includegraphics[width=\textwidth]{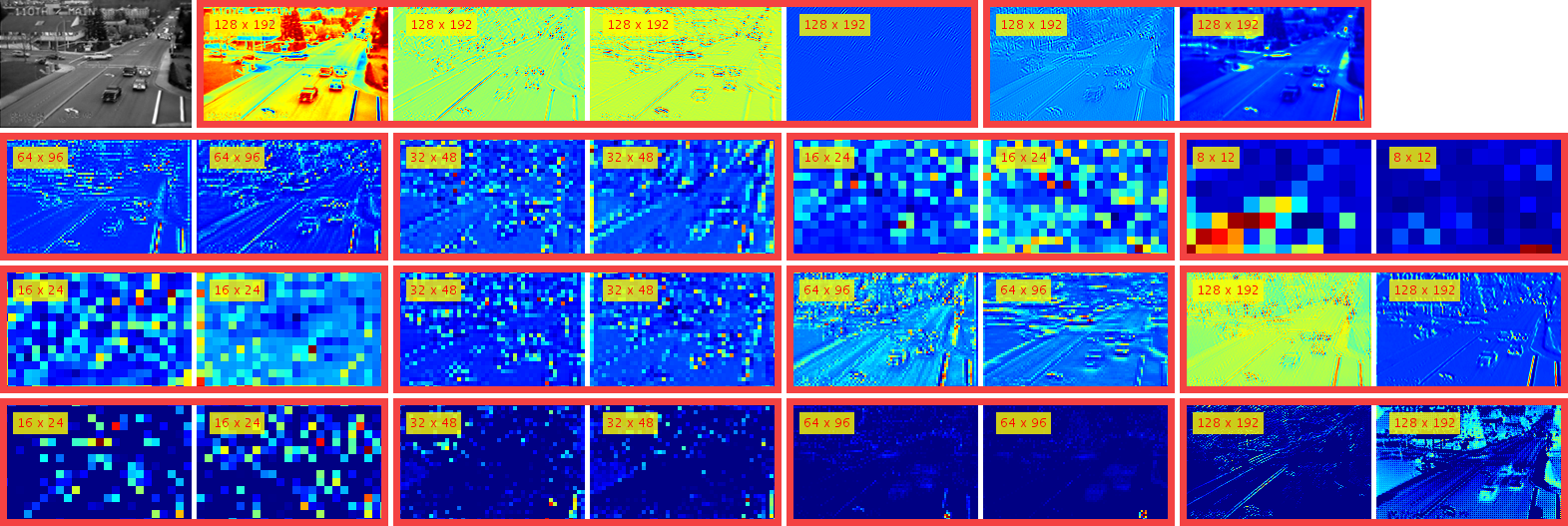}
\caption{Traffic-Belleview}
\end{subfigure}
\begin{subfigure}[t]{\textwidth}
\vspace{0.7cm}
\includegraphics[width=\textwidth]{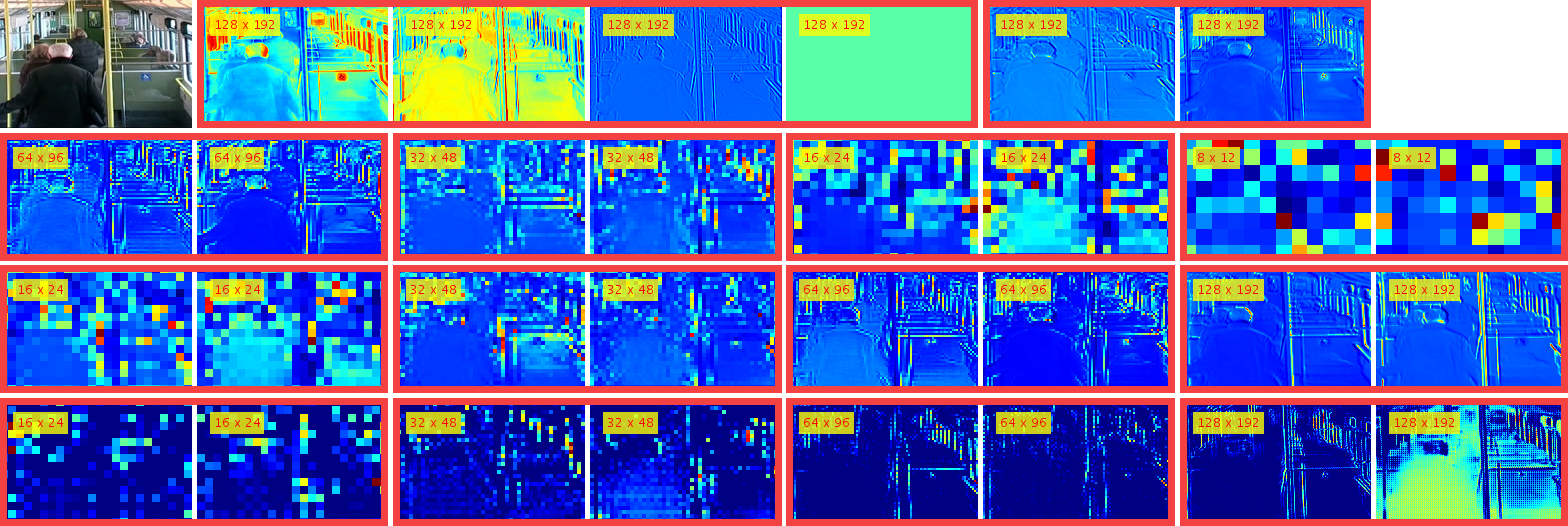}
\caption{Traffic-Train}
\end{subfigure}
\caption{Visualization of some activation maps (together with their spatial resolution) given an input frame for each dataset. Channels sampled from the same block are grouped by a red bounding box. Best viewed in color.}
\label{fig:feature_maps}
\end{figure*}

\section{Model optimization during training phase}
In this section, we show the outputs of the proposed model after some training epochs given the same input for each dataset. The number of training epochs and batch size are presented in Table~\ref{table:training}.
\begin{table}
\begin{center}
	\begin{tabularx}{\columnwidth}{ |l| *{2}{Y|} }%{|l|c|c|}
	\hline
	Dataset & Training epoch & Batch size \\
	\hline\hline
	UCSD Ped2 & 15 & 16 \\
	CUHK Avenue & 25 & 8 \\
	Subway Entrance & 25 & 16 \\
	Subway Exit & 15 & 8 \\
	Traffic-Train & 25 & 16 \\
	Traffic-Belleview & 120 & 8\\
	\hline
	\end{tabularx}
\end{center}
\caption{Number of training epochs and batch size in our experiments. These values were selected according to the number of training images in each dataset and the memory capacity of our hardware (Intel i7-7700K, 16 GB memory, GTX 1080).}
\label{table:training}
\end{table}

%\begin{figure}[t]
%\begin{center}
%{(a) UCSD Ped2}
%\includegraphics[width=\columnwidth]{figures/epochs/UCSDped2.png}
%{(b) CUHK Avenue}
%\includegraphics[width=\columnwidth]{figures/epochs/Avenue.png}
%{(c) Subway Entrance}
%\includegraphics[width=\columnwidth]{figures/epochs/Entrance.png}
%{(d) Subway Exit}
%\includegraphics[width=\columnwidth]{figures/epochs/Exit.png}
%{(e) Traffic-Belleview}
%\includegraphics[width=\columnwidth]{figures/epochs/Belleview.png}
%{(f) Traffic-Train}
%\includegraphics[width=\columnwidth]{figures/epochs/Train.png}
%\end{center}
%\caption{Visualization of model outputs provided by the two streams after some training epochs. Note that these input frames were from the test set and were not employed for training the models. Best viewed in color.}
%\label{fig:epoch_outputs}
%\end{figure}
\begin{figure*}[b]
\begin{subfigure}[t]{\textwidth}
\includegraphics[width=\textwidth]{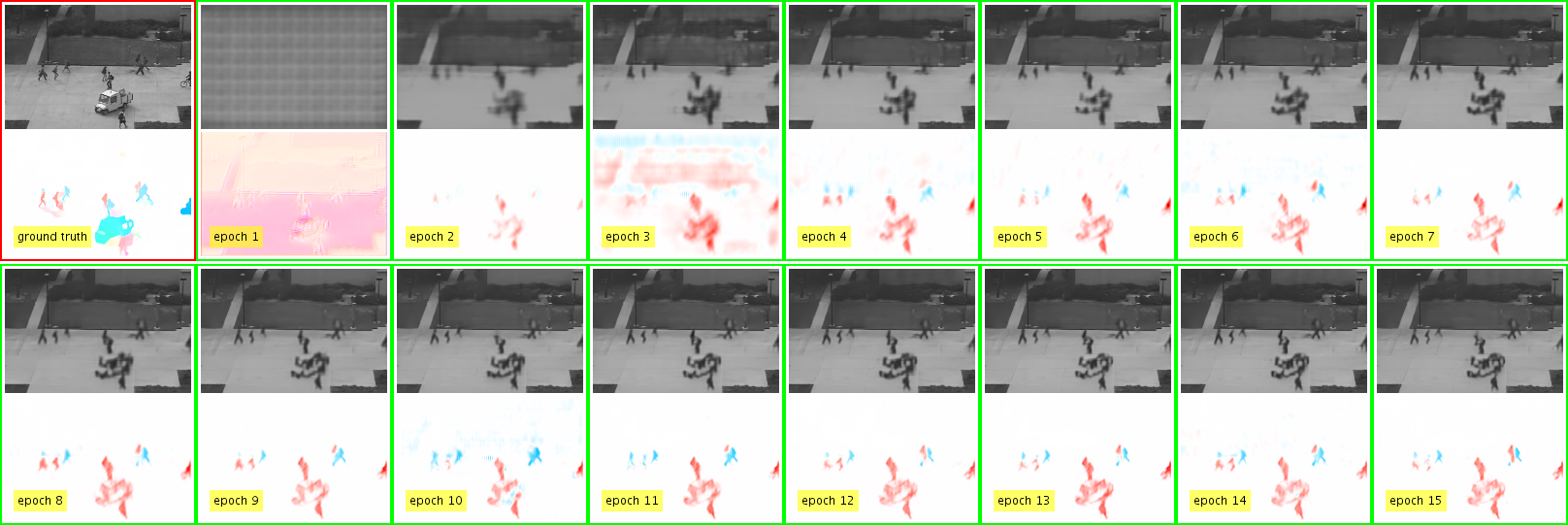}
\caption{UCSD Ped2}
\end{subfigure}
\begin{subfigure}[t]{\textwidth}
\vspace{1cm}
\includegraphics[width=\textwidth]{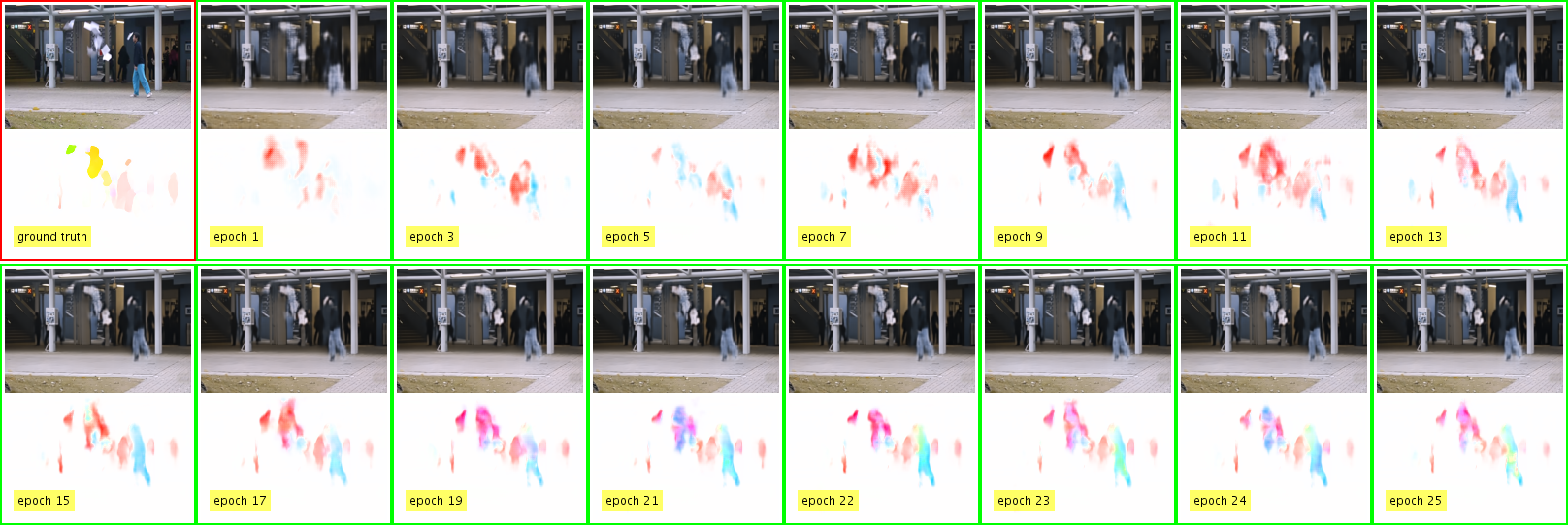}
\caption{CUHK Avenue}
\end{subfigure}
\begin{subfigure}[t]{\textwidth}
\vspace{1cm}
\includegraphics[width=\textwidth]{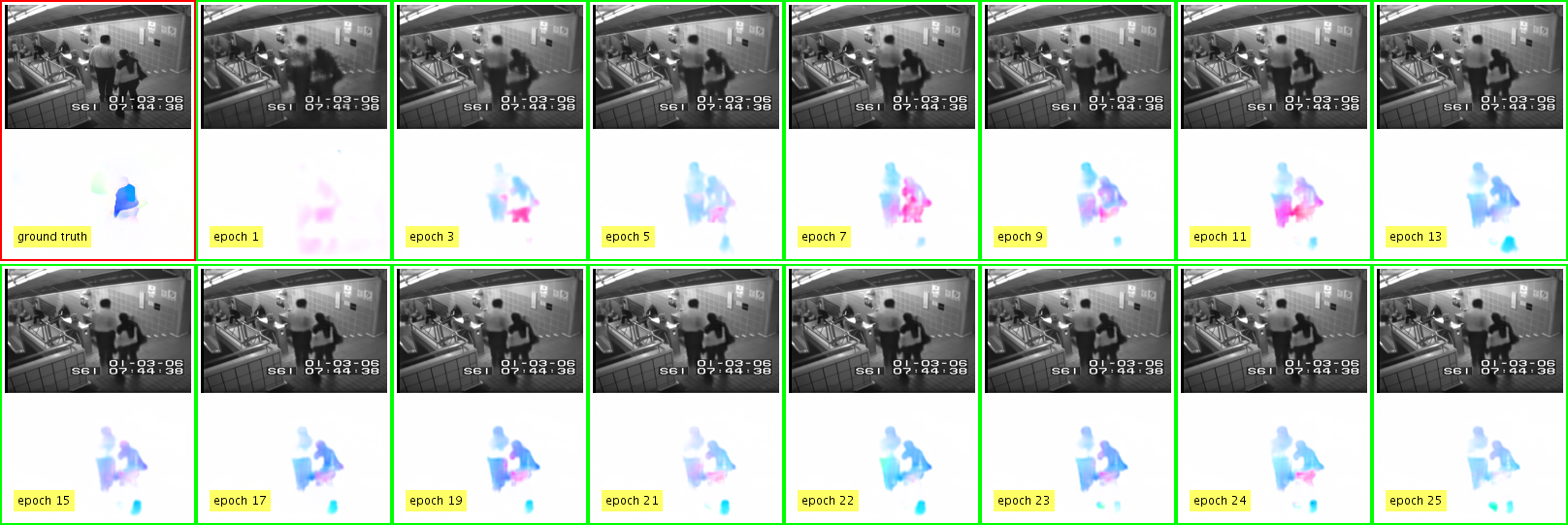}
\caption{Subway Entrance}
\end{subfigure}
\end{figure*}
\begin{figure*}\ContinuedFloat
\begin{subfigure}[t]{\textwidth}
\includegraphics[width=\textwidth]{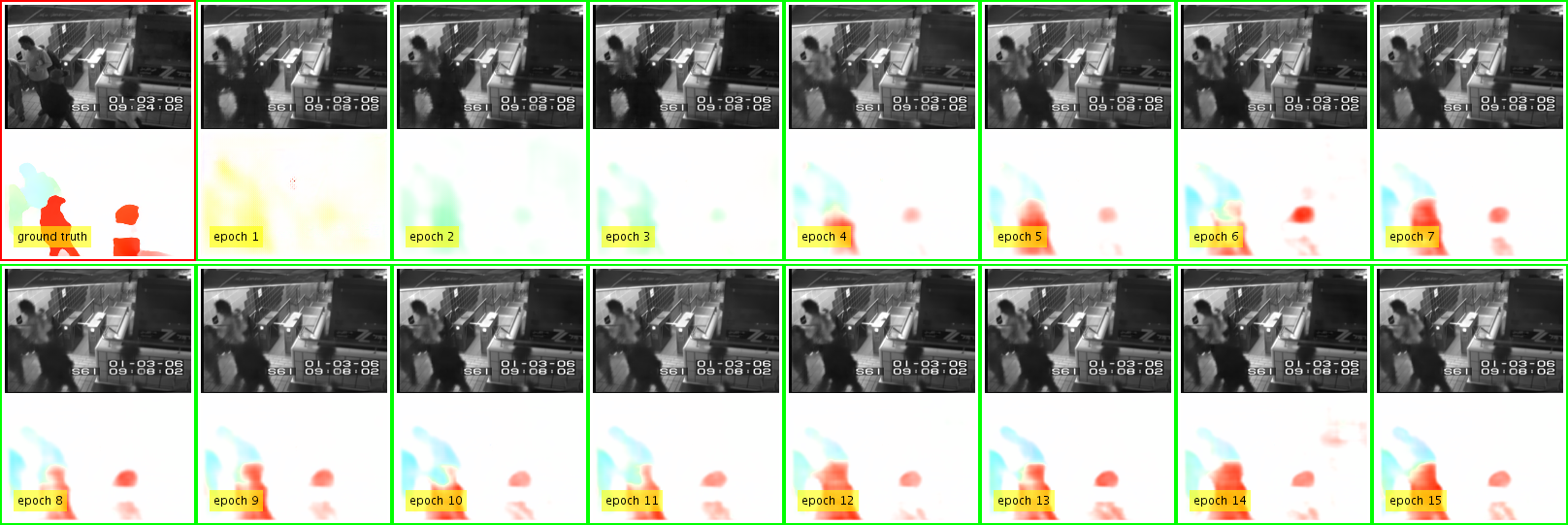}
\caption{Subway Exit}
\end{subfigure}
\begin{subfigure}[t]{\textwidth}
\vspace{1cm}
\includegraphics[width=\textwidth]{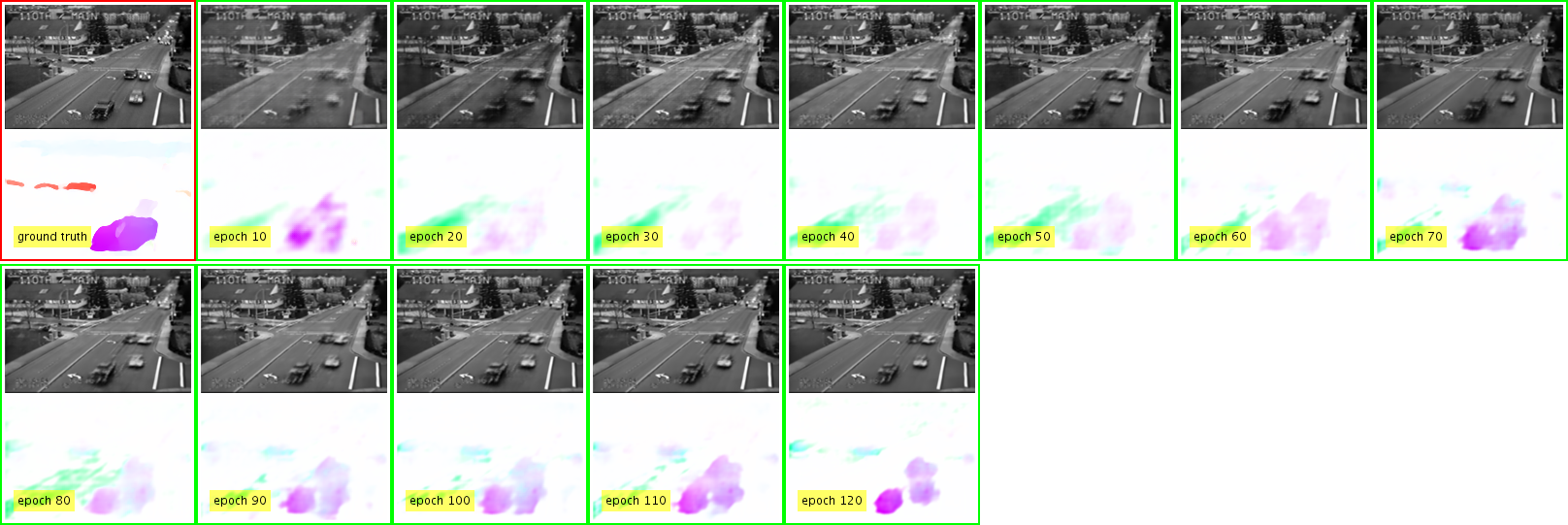}
\caption{Traffic-Belleview}
\end{subfigure}
\begin{subfigure}[t]{\textwidth}
\vspace{1cm}
\includegraphics[width=\textwidth]{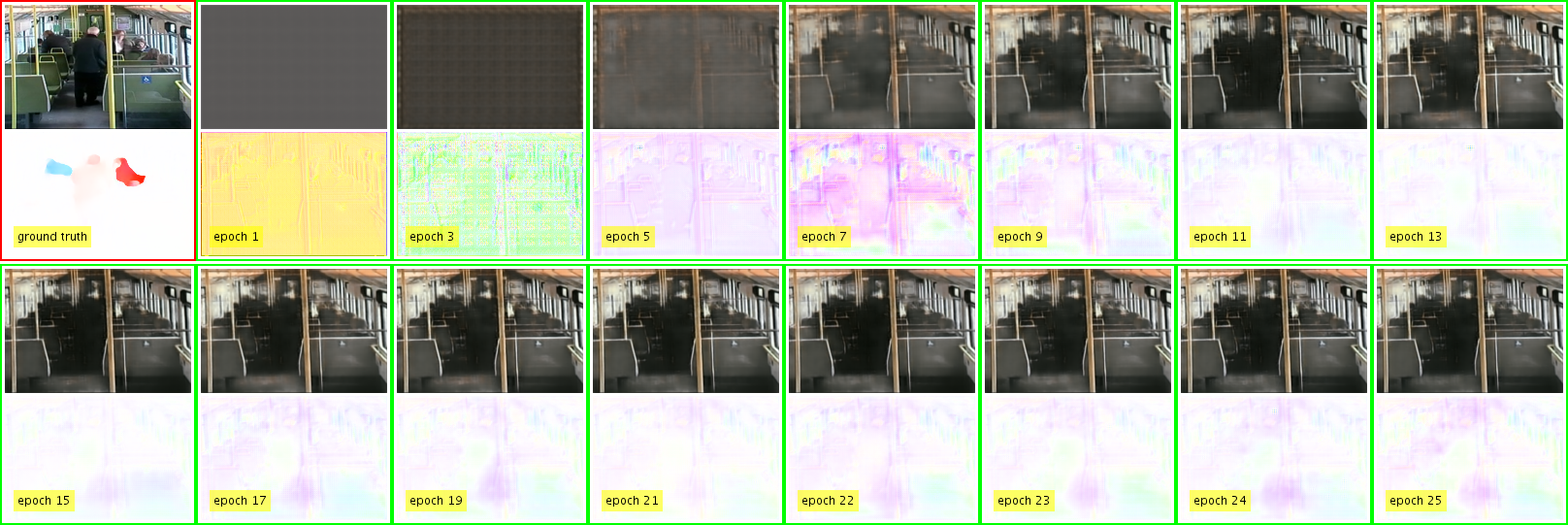}
\caption{Traffic-Train}
\end{subfigure}
\caption{Visualization of model outputs provided by the two streams after some training epochs. Note that these input frames were from the test set and were not employed for training the models. Best viewed in color.}
\label{fig:epoch_outputs}
\end{figure*}

In Figure~\ref{fig:epoch_outputs}, the correspondence between a reconstructed frame and its predicted motion can be clearly observed. A sharper frame would be obtained together with a motion with more details (\eg epochs 2 vs. 4 in the UCSD Ped2 experiment) as the number of epochs increases. It also demonstrates that the model encountered difficulty in optimizing the two streams on the Traffic-Train dataset due to the sudden change of lighting and camera jitter. However, the overall structure of the acquired scene was still preserved (\eg poles and seats). The use of SSIM on the input frame and its reconstruction hence improved the anomaly detection results (presented in the main paper).

{\small
\bibliographystyle{ieee_fullname}
\balance
\bibliography{egbib}
}